\documentclass{article} % For LaTeX2e
\usepackage{iclr2025_conference,times}
\usepackage{xspace}
\usepackage{colortbl,xcolor}
\usepackage{amsmath,amsfonts,bm}

\def\eqref#1{equation~\ref{#1}}
\def\1{\bm{1}}

\DeclareMathAlphabet{\mathsfit}{\encodingdefault}{\sfdefault}{m}{sl}
\SetMathAlphabet{\mathsfit}{bold}{\encodingdefault}{\sfdefault}{bx}{n}

\definecolor{pearDark}{HTML}{2980B9}
\definecolor{mycolor_green}{HTML}{E6F8E0}
\usepackage{booktabs} 
\usepackage{multirow}
\usepackage{hyperref}
\usepackage{url}
\usepackage{graphicx}
\usepackage{marvosym}

\newcommand{\methodname}{Bootstrap3D\xspace}

\title{\methodname: Improving Multi-view Diffusion Model with Synthetic Data}

\author{
Zeyi Sun$^{1,3}$, Tong Wu$^{2}${\textsuperscript{\Letter}}, Pan Zhang$^{3}$, Yuhang Zang$^{3}$, Xiaoyi Dong$^{3}$ \\
\textbf{Yuanjun Xiong$^{4}$, Dahua Lin$^{2,3}$, Jiaqi Wang$^{3}${\textsuperscript{\Letter}}} \vspace{2mm}\\
$^1$Shanghai Jiao Tong University \quad 
$^2$The Chinese University of Hong Kong \quad  \\
$^3$Shanghai AI Laboratory \quad $^4$MThreads, Inc.\\
{{\centering Project Page: \url{https://SunzeY.github.io/Bootstrap3D/}}
\vspace{-6mm}
}
}

\iclrfinalcopy % Uncomment for camera-ready version, but NOT for submission.
\begin{document}

\maketitle

\begin{abstract}
Recent years have witnessed remarkable progress in multi-view diffusion models for 3D content creation. However, there remains a significant gap in image quality and prompt-following ability compared to 2D diffusion models. A critical bottleneck is the scarcity of high-quality 3D data with detailed captions. To address this challenge, we propose \textbf{\methodname}, a novel framework that automatically generates an arbitrary quantity of multi-view images to assist in training multi-view diffusion models. Specifically, we introduce a data generation pipeline that employs (1) 2D and video diffusion models to generate multi-view images based on constructed text prompts, and (2) our fine-tuned 3D-aware \textbf{MV-LLaVA} for filtering high-quality data and rewriting inaccurate captions. Leveraging this pipeline, we have generated 1 million high-quality synthetic multi-view images with dense descriptive captions to address the shortage of high-quality 3D data. Furthermore, we present a \textbf{Training Timestep Reschedule (TTR)} strategy that leverages the denoising process to learn multi-view consistency while maintaining the original 2D diffusion prior. Extensive experiments demonstrate that \methodname can generate high-quality multi-view images with superior aesthetic quality, image-text alignment, and maintained view consistency.
\end{abstract}

\section{Introduction}

3D content creation stands as a fundamental challenge within the generative domain, boasting widespread applications in augmented reality (AR) and game modeling. Unlike 2D image generation, the dearth of high-quality 3D models persists as a significant hurdle to overcome. In the realm of 2D image generation, the pivotal role of training on billion-scale image-text pairs~\citep{schuhmann2022laion} has been firmly established~\citep{betker2023improving,rombach2022high,li2024playground,chen2023pixart,chen2024pixart}. However, in 3D content generation, the scarcity of high-quality 3D models often compels reliance on the priors of 2D diffusion models. The predominant methodologies in this domain can be categorized into two main streams: 1) Gaining optimized neural representations from fixed 2D diffusion models via Score Distillation Sampling (SDS) loss~\citep{poole2022dreamfusion,shi2023mvdream,liu2023zero,shi2023zero123++,liu2023one,wang2024prolificdreamer}, which are time-intensive, lacking diversity and suffer from low robustness although capable of producing high-quality 3D objects.
% \tong{and suffer from a low robustness (add some more disadvantages)}. 
2) Fine-tuning 2D diffusion models to achieve multi-view generation~\citep{li2023instant3d,shi2023zero123++,shi2023mvdream}
% \tong{not sure nvs is a proper representation, text-to-4view can not be named as nvs?}
, directly synthesizing 3D objects through sparse reconstruction models~\citep{li2023instant3d,wang2023pflrm,xu2024instantmesh,xu2024grm,tang2024lgm,wei2024meshlrm}. With recent improvements in large-scale sparse view reconstruction models and 3D representations~\citep{kerbl20233d}, the second stream is garnering increasing attention. 
% \tong{1) The grammar of this long sentence is a little bit strange. 2) The ending focus on sparse-view recon, while it should be sparse-view gen considering the content of the paper.}

Fine-tuning 2D diffusion models for multi-view generation remains challenging owing to the insufficiency in both data quality and quantity. Previous methods~\citep{qiu2023richdreamer,li2023instant3d,shi2023mvdream,deitke2024objaverse} primarily train on a filtered subset of high-quality data from Objaverse~\citep{deitke2023objaverse} and Objaverse-XL~\citep{deitke2024objaverse}. The scarcity of high-quality data often introduces various shortcomings. In single-view based novel view synthesis~\citep{liu2023zero,shi2023zero123++,wang2023imagedream,voleti2024sv3d}, if the input images deviate from the distribution of the training data, it can induce issues such as motion blurring, object distortion and deformation~\citep{shi2023zero123++}.

\begin{figure}[t]
\begin{center}
\includegraphics[width=1.0\textwidth]{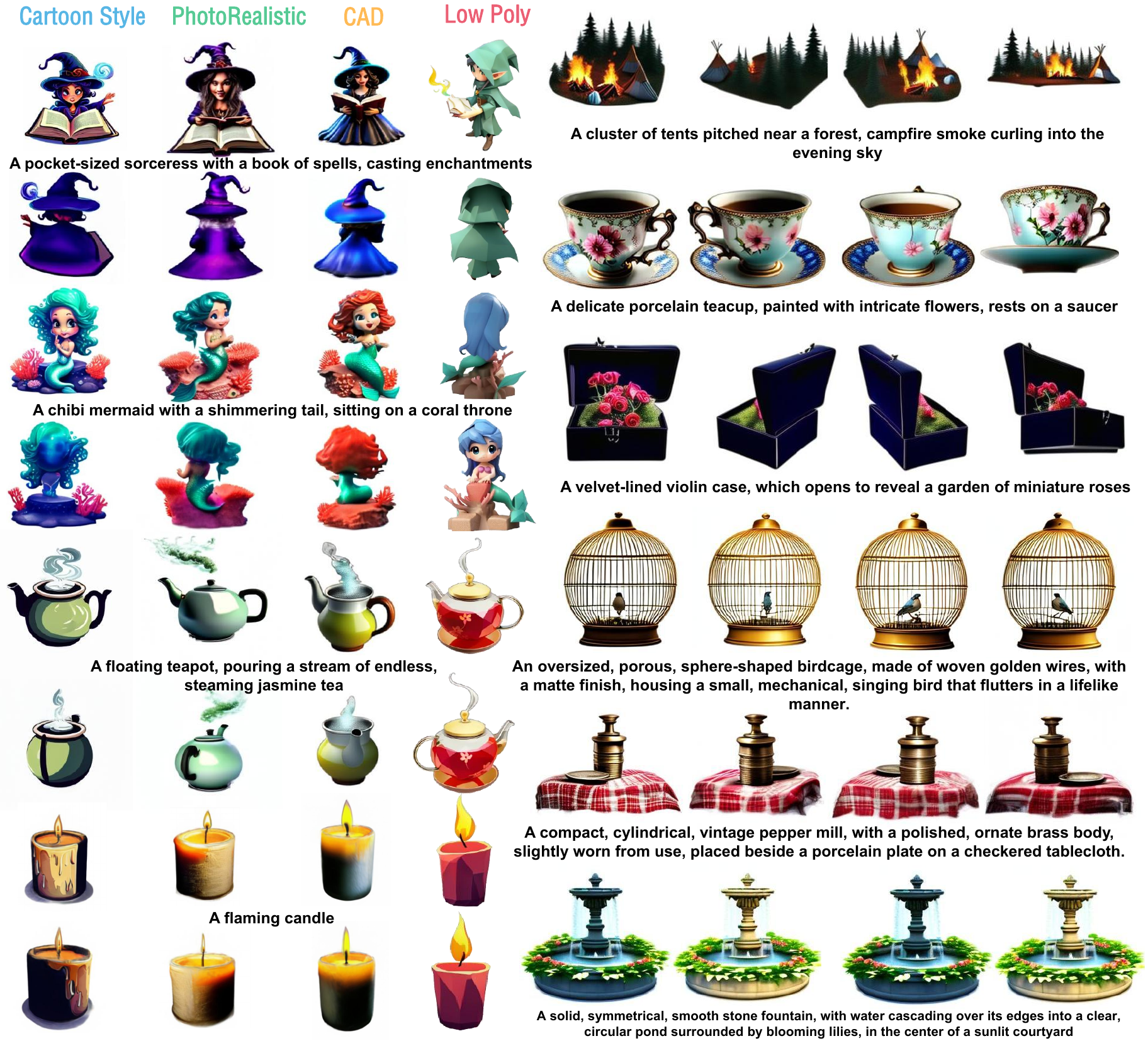}
\vspace{-7mm}
    \caption{\textbf{Bootstrap3D} can generate high quality multi-view images with precise long text control and style customization while maintaining view consistency.}
    \label{fig:teaser}
\vspace{-10mm}
\end{center}
\end{figure}

Moreover, in direct text-to-multi-view image generation, the pursuit of enhancing view consistency compromises the aesthetic and photo-realistic quality. For instance, Intant3D~\citep{li2023instant3d} fine-tunes SDXL~\citep{podell2023sdxl} using only 10K high-quality Objaverse~\citep{deitke2023objaverse} data with a small learning rate for 10K steps, which does not fundamentally prevent the catastrophic forgetting problem of losing 2D diffusion prior, leading to compromised image quality. Recent endeavors have predominantly focused on alleviating data scarcity and improving view consistency from a model-centric perspective~\citep{kant2024spad,shi2023zero123++,tang2024mvdiffusion++}, with limited exploration into the improvement of training data itself.

Recent Multimodal Large Language Models (MLLMs)~\citep{liu2024visual,chen2023sharegpt4v,Li2023BLIP2BL,Alayrac2022FlamingoAV,Anil2023PaLM2T} like GPT-4V~\citep{2023GPT4VisionSC} and Gemini~\citep{team2023gemini}, possess image understanding capabilities and rudimentary 3D world awareness, has enabled automatic quality assessment of multi-view images and dense caption generation. Furthermore, notable advancements in video diffusion~\citep{videoworldsimulators2024,voleti2024sv3d} have improved the generalizability of novel view synthesis~\citep{voleti2024sv3d,chen2024v3d,kwak2023vivid}. Employing these advancements, we propose \methodname to generate synthetic data to counteract the data deficiencies inherent in training multi-view diffusion models.
To be specific, we introduce the \methodname data generation pipeline for producing high-quality multi-view images with dense descriptive captions. Subsequently, we fine-tune a multi-view-aware MLLM model, dubbed as MV-LLaVA, to achieve fully automated high-quality data annotation with both efficiency and accuracy. To mitigate catastrophic forgetting of 2D diffusion prior, we introduce a training timestep reschedule (TTR) strategy when fine-tuning multi-view diffusion models. Specifically, we use the phased nature of the denoising process~\citep{ho2020denoising} and limit different training time steps for synthetic data to achieve enhanced image quality with maintained view consistency.
% We also propose a method to leverage the phased nature of the denoising process~\citep{ho2020denoising} by limiting training time steps for different data types to mitigate the catastrophic forgetting of 2D diffusion priors. 
% \tong{The introduction of method lacks details. The background and motivation here (MLLM and SVD) is even longer than the content of the concrete method.}

% \tong{Make the experimental observations and achievements a separate paragraph}
Through extensive experiments, we demonstrate that our method significantly enhances the adherence of the multi-view diffusion model to text prompts and image quality while ensuring view consistency. Integrated with the reconstruction model, our approach facilitates the creation of 3D models with superior quality. Our contributions are summarized into the following points:

\textbf{1)} We present an automated \textbf{\methodname} data generation pipeline that uses the video diffusion model and our fine-tuned 3D-aware MV-LLaVA to synthesize an arbitrary number of high-quality multi-view image text pairs. 
% We propose a data generation pipeline for producing high-quality multiview images with dense descriptive captions. Leveraging a fine-tuned multi-view-aware MLLM model MV-LLaVA, we achieve fully automated data generation with both efficiency and accuracy.

\textbf{2)} We propose a Training Time-step Reschedule (TTR) strategy for fine-tuning the multi-view diffusion model that employs both synthetic data and real data to enhance image quality and image-text alignment while maintaining view consistency.

\textbf{3)} We generate \textbf{1 million} multi-view images with dense descriptive captions suitable for training the multi-view diffusion model, which improves the 3D generation quality and mitigates the gap with the 2D diffusion model from a data perspective.
% . Our experiments validate that this strategy can significantly enhance the quality of generated multi-view images and the model's ability to faithfully follow text prompts while maintaining view consistency.

% \tong{some details in the contribution section should be move above, make this more concise}
\section{Related Work}
\textbf{Existing 3D datasets and data pre-processing.} Existing object level 3D datasets, sourced either from CAD~\citep{chang2015shapenet,wu20153d,deitke2023objaverse,deitke2024objaverse} or scan from real objects~\citep{aanaes2016large,yao2020blendedmvs,downs2022google,wu2023omniobject3d}, are still small in size. Most state-of-the-art open-sourced 3D content creation models are trained on Objaverse~\citep{deitke2023objaverse}. However, there still exists a huge gap compared to data used for training 2D diffusion models~\citep{schuhmann2022laion}. In addition to quantity, quality is also an important problem remains to be solved as many methods~\citep{shi2023mvdream,li2023instant3d,qiu2023richdreamer,tang2024lgm} trained on Objaverse rely on filtering out low-quality data, making the precious 3D data even less.
% This gap poses a significant obstacle to achieving high-quality 3D content creation compared to its 2D diffusion counterpart. 
Another critical gap that requires attention is the quality of the 3D object's caption. Previous work Cap3D~\citep{luo2024scalable} propose to apply BLIP-2~\citep{Li2023BLIP2BL} and GPT-4~\citep{openai2023gpt} to generate caption based on multi-view images. However, this approach,  without direct input image into GPT, can lead to severe hallucination. Given recent breakthroughs in improving text-image alignment through caption rewriting~\citep{betker2023improving,chen2023pixart,chen2024pixart,esser2024scaling}, there is a pressing need to rewrite denser and more accurate captions for 3D objects with the assistance of advanced Multimodal Large Language Models (MLLMs). In this work, we propose a new data generation pipeline to synthesize multi-view images and rewrite captions for 3D objects incorporating additional quality scoring mechanisms to address the aforementioned issues.

\textbf{Text-to-3D content creation.}
The field of 3D content creation has been a vibrant area of research over the past years. One prominent research direction explores the use of Score Distillation Sampling (SDS)~\citep{poole2022dreamfusion} and its variants~\citep{chen2023fantasia3d,chung2023luciddreamer,hertz2023delta,liang2023luciddreamer,lin2023magic3d,liu2023zero,shi2023mvdream,liu2023syncdreamer,long2023wonder3d,wang2024prolificdreamer,tang2023dreamgaussian,wang2023score,yang2024magic,qi2024tailor3d}, using the priors of 2D diffusion models to optimize 3D representations. While these methods have demonstrated success in producing high-quality 3D generations, they often require prolonged optimization time to converge. In contrast, recent studies~\citep{hong2023lrm,wang2023pflrm,li2023instant3d,tang2024lgm,tochilkin2024triposr,xu2024grm,wei2024meshlrm} have proposed the direct inference of 3D representations~\citep{mildenhall2021nerf,chan2022efficient,kerbl20233d,zhang20233dshape2vecset} conditioned by images. Among these approaches, Instant3D~\citep{li2023instant3d} stands out by utilizing multi-view images of the same object to directly deduce the Triplane~\citep{chan2022efficient} representation. This approach effectively addresses the issue of ambiguous unseen areas inherent in the single image to 3D conversions, as encountered in LRM~\citep{hong2023lrm} and TripoSR~\citep{tochilkin2024triposr}. Instant3D, along with subsequent works~\citep{xu2024grm,zheng2024mvd,wang2024crm,xu2024instantmesh}, efficiently decomposes 3D generation into two processes: the generation of multi-view images using multi-view diffusion model~\citep{liu2023zero,liu2023syncdreamer,liu2023one,shi2023mvdream,liu2024one,shi2023zero123++,long2023wonder3d,kant2024spad,voleti2024sv3d} and large reconstruction model to generate 3D representations conditioned on these multi-view images. In this work, we introduce a method that significantly enhances the scalability of training and data generation for multi-view image generation. 
% Our approach involves training a multi-view diffusion model on vast, curated synthetic data, which substantially improves the controllability of text prompts and the quality of generated objects through the synergistic integration of sparse view reconstruction models like Instant3D~\citep{li2023instant3d} and GRM~\citep{xu2024grm}.

\textbf{Video diffusion for novel view synthesis.}
Recent advancements in video diffusion have marked a significant breakthrough, with models such as Sora~\citep{videoworldsimulators2024} and SVD~\citep{blattmann2023stable} scaling up the direct generation process from images to videos. Following these developments, a series of works~\citep{wang2023motionctrl,kwak2023vivid,blattmann2023stable,melas20243d,han2024vfusion3d,chen2024v3d} represented by SV3D~\citep{voleti2024sv3d}, have fine-tuned these video diffusion models for 3D content creation. Despite these groundbreaking developments, the new perspective images generated based on video priors still suffer from issues like motion blur. In this work, we propose to utilize SV3D~\citep{voleti2024sv3d} as a data generator to produce novel views of given images with additional quality checks to leave only high-quality data.

\textbf{Multimodal Large Language Models.} With the development of large language models~\citep{Brown2020LanguageMA,openai2023gpt,Chowdhery2022PaLMSL,Anil2023PaLM2T,Hoffmann2022TrainingCL,Touvron2023LLaMAOA}, multimodal large language models (MLLMs)~\citep{zhang2023internlmxcomposer,Alayrac2022FlamingoAV,Li2023BLIP2BL,Li2022BLIPBL,Huang2023LanguageIN,Driess2023PaLMEAE,Awadalla2023OpenFlamingoAO,liu2024visual,dong2024internlmxcomposer2,sun2023alphaclip}, such as GPT-4V~\citep{2023GPT4VisionSC}, have demonstrated groundbreaking 2D comprehension capabilities and open-world knowledge. As is discovered in GPTEval3D~\citep{wu2024gpt}, GPT-4V can achieve human-aligned evaluation for multi-view images rendered from 3D objects. In this work, we fine-tune the LLaVA~\citep{liu2024visual}  for quality judgment and descriptive caption generation based on multi-view images.
\vspace{-2mm}

\begin{figure*}[t]
  \centering
  \includegraphics[width=1.0\linewidth]{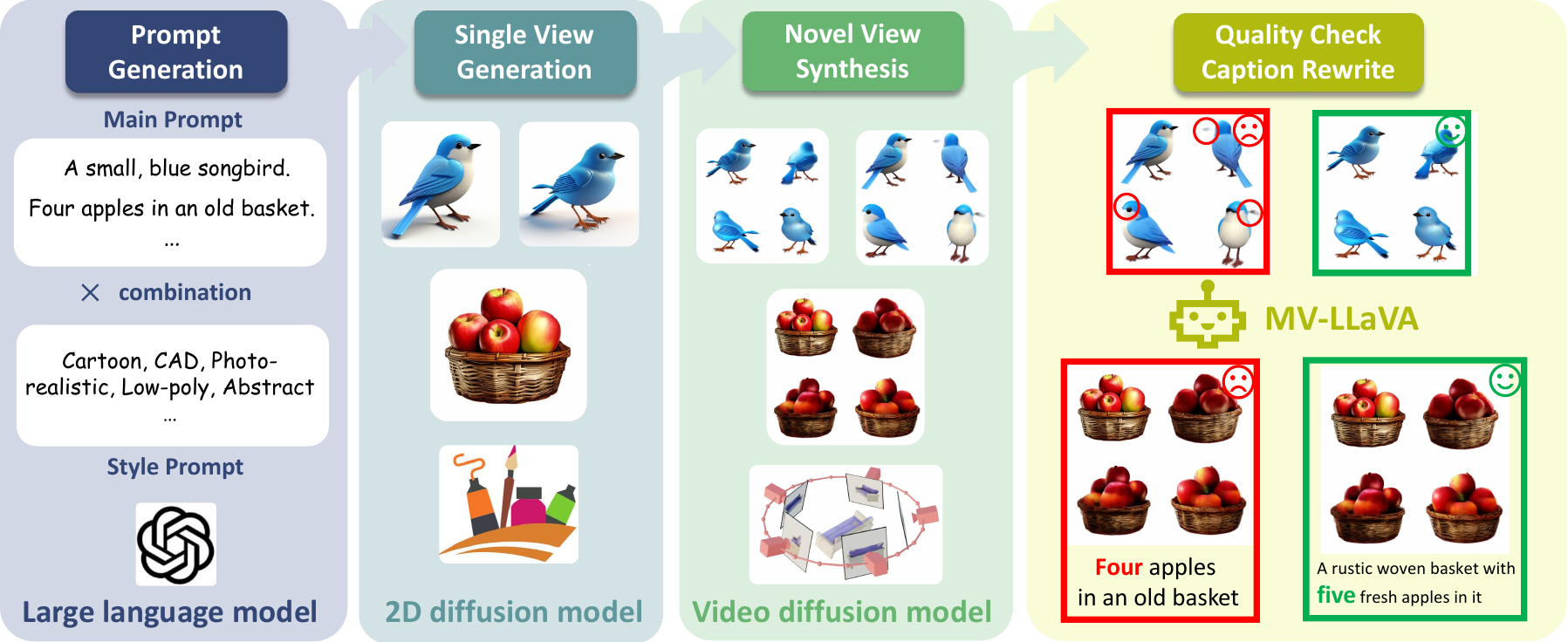}   \vspace{-6mm}
    \caption{\textbf{\methodname data generation pipeline} that consists of 1) using LLM to generate diverse text prompts 2) employing the T2I model to generate single-view images 3) synthesizing arbitrary number of multi-view images by applying the video diffusion model, 4) employing MV-LLaVA to filter and select only high-quality data, and rewrite captions to be dense and descriptive.}
  \vspace{-6mm}
  \label{fig:data_pipeline}
\end{figure*}

\section{Methods}
Due to the scarcity of high-quality 3D data, we develop the \methodname data generation pipeline to efficiently construct an arbitrary number of training data (Sec. \ref{sec:data_generation}).
Subsequently, the quality of generated multi-view images is assessed using the powerful GPT-4V~\citep{2023GPT4VisionSC} or our proposed MV-LLaVA~\citep{liu2024visual} model to generate dense descriptive captions efficiency and faithfully (Sec. \ref{sec:mv_llava}). We also design a training timestep reschedule (Sec. \ref{sec:ttr}) when fine-tuning the multi-view diffusion model with our synthetic and real data.

\subsection{\methodname Data Generation Pipeline}\label{sec:data_generation}
As illustrated in Fig.\ref{fig:data_pipeline}, our data generation pipeline initially employs GPT-4~\citep{2023GPT4VisionSC} to generate a multitude of imaginative and varied text prompts~\citep{wu2024gpt}. Subsequently, to generate 2D images that closely align with the text prompts, we utilize the PixArt-Alpha~\citep{chen2023pixart} model use FlanT5~\citep{chung2024scaling} text encoder with DiT~\citep{peebles2023scalable} architecture for text-to-image (T2I) generation. Thereafter, we use SV3D~\citep{voleti2024sv3d} for novel view synthesis. Given the significant motion blur and distortion often present in SV3D~\citep{voleti2024sv3d} outputs, we further employ Multimodal Large Language Models(MLLM) to evaluate the quality of multi-view images. To rectify mismatches between multi-view images and the original text prompts induced by novel view synthesis and provide more precise captions, we further propose MV-LLaVA to generate dense descriptive captions for multi-view images.
% Utilizing the aforementioned data pipeline, we constructs 560K multi-view images with corresponding dense captions. Through our newly trained MLLM, we filter out 220K high-quality data for subsequent training.

\subsection{Multi-View LLaVA (MV-LLaVA)}\label{sec:mv_llava}
To efficiently generate captions and label quality scores for both generated multi-view images and 3D assets in Objaverse~\citep{deitke2023objaverse}, we propose the Multi-View LLaVA (MV-LLaVA) that fine-tune LLaVA~\citep{liu2024visual} based on our instructive conversation pairs generated by the powerful GPT-4V~\citep{2023GPT4VisionSC}.

\noindent \textbf{Preparing the instruction tuning data.} As shown in Fig.\ref{fig:data_pipeline}, we use GPT-4 to generate 20k varied text prompts based on prompts designed in~\citep{wu2024gpt} and use PixArt-alpha~\citep{chen2023pixart} to generate single view image and use SV3D~\citep{voleti2024sv3d} or Zero123++~\citep{shi2023zero123++} to generate multi-view images. For these 20k generated multi-view images, we prompt GPT-4V~\citep{2023GPT4VisionSC} to generate comments on view consistency, image quality and generate dense descriptive captions. For the additional 10K rendered multi-view images from Objaverse~\citep{deitke2023objaverse}, we prompt GPT-4V (detailed prompts in Sup.~\ref{prompt_detail}) to offer feedback on the quality and aesthetic appeal of 3D objects, along with style judgments. We utilize these 30K high-quality multi-view image text pairs (prompts detailed in Sup.~\ref{sup:llava_prompt_detail}) as the instruction tuning data for LLaVA.

\begin{figure*}[t]
  \centering
  \includegraphics[width=1.0\linewidth]{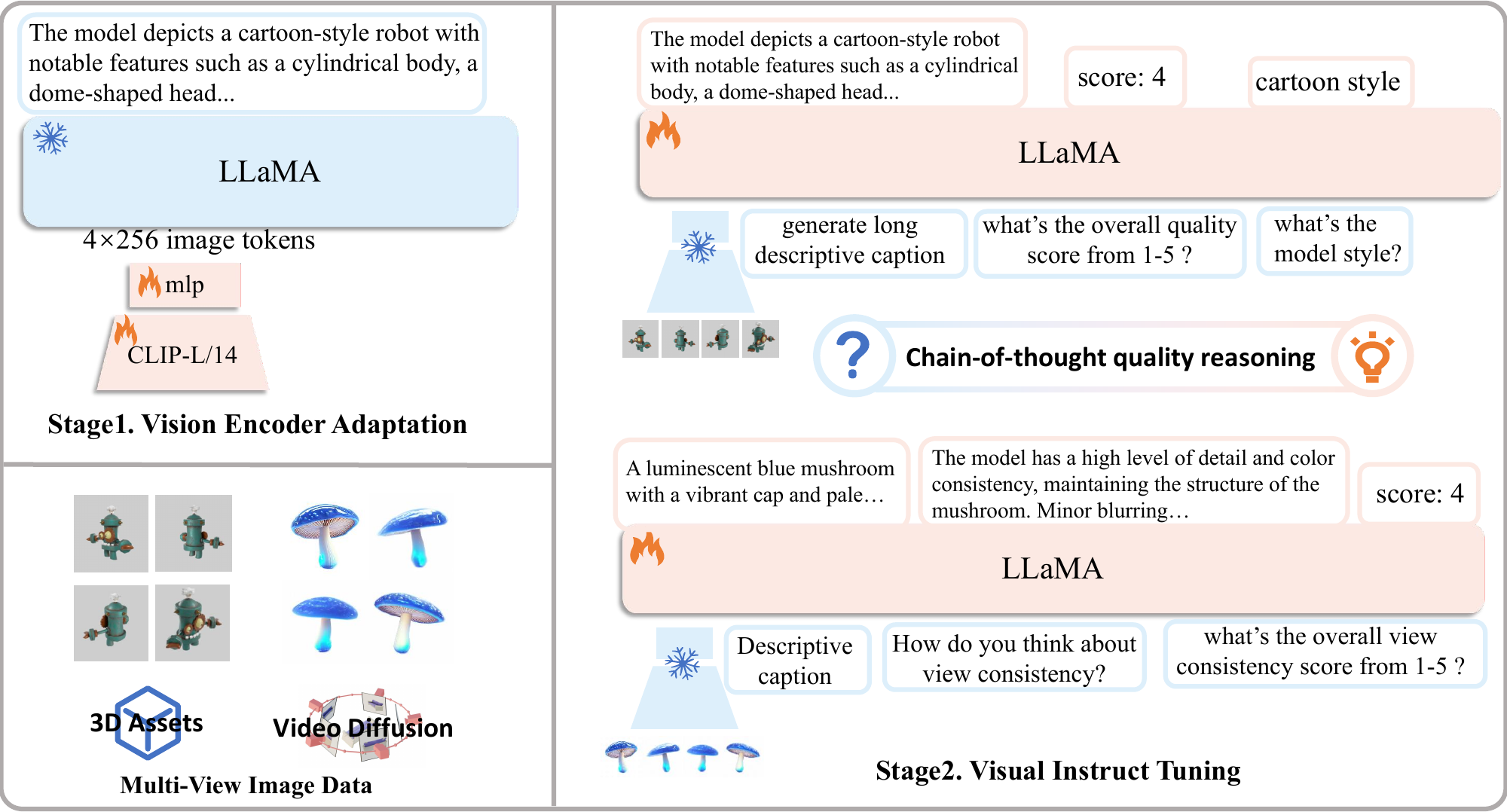}
  \caption{\textbf{MV-LLaVA.} We use GPT-4V~\citep{2023GPT4VisionSC} to generate long descriptive captions, quality scoring, and reasoning processes for multi-view images to construct the instruction tuning dataset. Then we fine-tune our MV-LLaVA based on LLaVA~\citep{liu2024visual} to serve as the human-aligned quality checker and captioner for multi-view images.}
  \vspace{-6mm}
   \label{fig:mvllava}
\end{figure*}

\noindent \textbf{Instruction tuning.} As presented in the left part of Fig.~\ref{fig:mvllava}, due to the LLaVA's maximum training context length constraints of 2048, we input four images separately into CLIP-L/14~\citep{radford2021learning} and generate 4$\times$256 image tokens. Inspired by ShareGPT-4V~\citep{chen2023sharegpt4v}, we freeze only a portion of layers of CLIP~\citep{radford2021learning} in the first stage of pre-training to enhance multi-view awareness and texture perception of vision encoder (detailed in Sup.~\ref{sup:mvllava_layers}).
As shown in the right part of Fig.~\ref{fig:mvllava}, we first ask the model to generate descriptions, then let the model score the quality based on multi-view images and captions. Our approach encourages LLM to deduct more reasonable quality scores through chain-of-thought~\citep{wei2022chain}. A mixture of original training data of LLaVA is adopted to mitigate over-fitting. As a result, we obtain MV-LLaVA, which efficiently filters and re-captions both synthetic data and 3D assets. As detailed in Sup.\ref{sec:quality_of_mvllava}, MV-LLaVA can not only generate more accurate, less hallucinated dense captions that faithfully describe 3D objects compared to Cap3D~\citep{luo2024scalable} but also assign the human-aligned quality score on both synthetic data and Objaverse assets. The filtered high-quality multi-view images with rewritten dense captions served as training data for the diffusion model.

\subsection{Training Timestep Reschedule (TTR)}\label{sec:ttr}
\label{method_tnr}
\begin{figure*}[t]
  \centering
  \includegraphics[width=.95\linewidth]{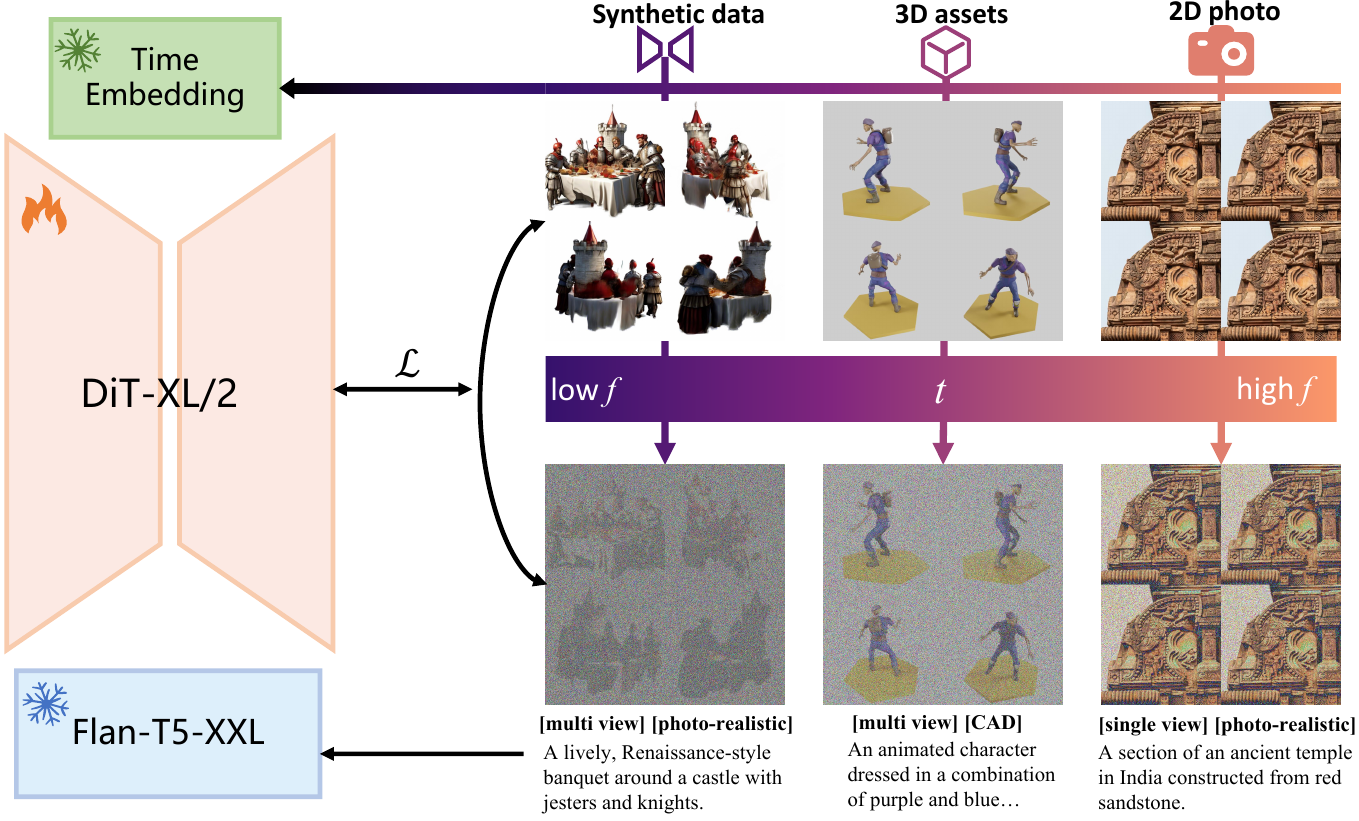}
  \vspace{-4mm}
  \caption{\textbf{Training Timestep Reschedule (TTR).} For different types of training data, we restrict the training time step $t$ accordingly to achieve the balance between varied high aesthetic images that are better aligned with text prompt, photo-realistic texture, and view consistency for 3D generation.}
  \vspace{-2mm}
  \label{fig:time_step}
\end{figure*}

Despite retaining only relatively high-quality synthetic data with minimal motion blur from SV3D~\citep{voleti2024sv3d} through MV-LLaVA, small areas of blurring persist, stemming from both motion and out-of-distribution scenarios for SV3D and SVD~\citep{blattmann2023stable}. These blurred data can potentially compromise the final performance of the multi-view diffusion model. To restrict the training time step for synthetic data, we proposed a simple yet effective Training Timestep Reschedule (TTR) method.

\noindent \textbf{Background.} Before delving into TTR, we briefly review some basic concepts needed to understand diffusion models (DDPMs)~\citep{ho2020denoising,sohl2015deep,salimans2022progressive,rombach2022high,chen2023pixart}. Gaussian diffusion models assume a forward noising process which
gradually applies noise to real data $x_0$
\begin{align}
q(x_t|x_0) = \mathcal{N}(x_t; \sqrt{\bar{\alpha}_t}x_0, (1 - \bar{\alpha}_t)\mathbf{I})
\end{align}
here constants $\bar{\alpha}_t$ are hyperparameters. By applying the reparameterization trick, we can sample
\begin{align}
x_t = \sqrt{\bar{\alpha}_t}x_0 + \sqrt{1 - \bar{\alpha}_t} \epsilon_t
\end{align}
During training, $t$ is randomly sampled in $[0, N]$ ($N=1000$ in ~\citep{chen2023pixart,rombach2022high}) for the model to predict the added noise $\epsilon_t$, where $x_0$ denotes for the clear nature image and $x_N$ denotes for pure Gaussian noise. As depicted in Fig.\ref{fig:time_step}, when $t$ is large, the denoising process primarily focuses on determining the global
low frequency($f$) content such as overall structure and shape. Conversely, when $t$ is small, the denoising process is predominantly responsible for generating high $f$
components such as texture. 

\noindent When adapting Stable Diffision~\citep{rombach2022high} for multi-view generation, the previous approach~\citep{shi2023zero123++} changes the default scaled linear schedule into the linear schedule to emphasize more on early denoising stage for structural variation and view consistency. Inspired by this, we propose restricting the denoising time step of synthetic data during training. As small yet observable blur still exists in synthetic data with novel view generated by SV3D~\citep{voleti2024sv3d}, we limit them to training diffusion model only with large $t$. This restricts the backpropagation of these synthetic data to focus on the low-frequency component of the image like the overall structure and shape that faithfully follow text prompts and consistency between different views. Small $t$ values are only sampled on clear and physically consistent multi-view images rendered from Objaverse~\citep{deitke2023objaverse} and supplemented high-quality 2D images from SA-1B~\citep{kirillov2023segment}, help model outcome high-quality images with more photo-realistic and varied texture details.
\begin{figure*}[t]
  \centering
  \includegraphics[width=1.0\linewidth]{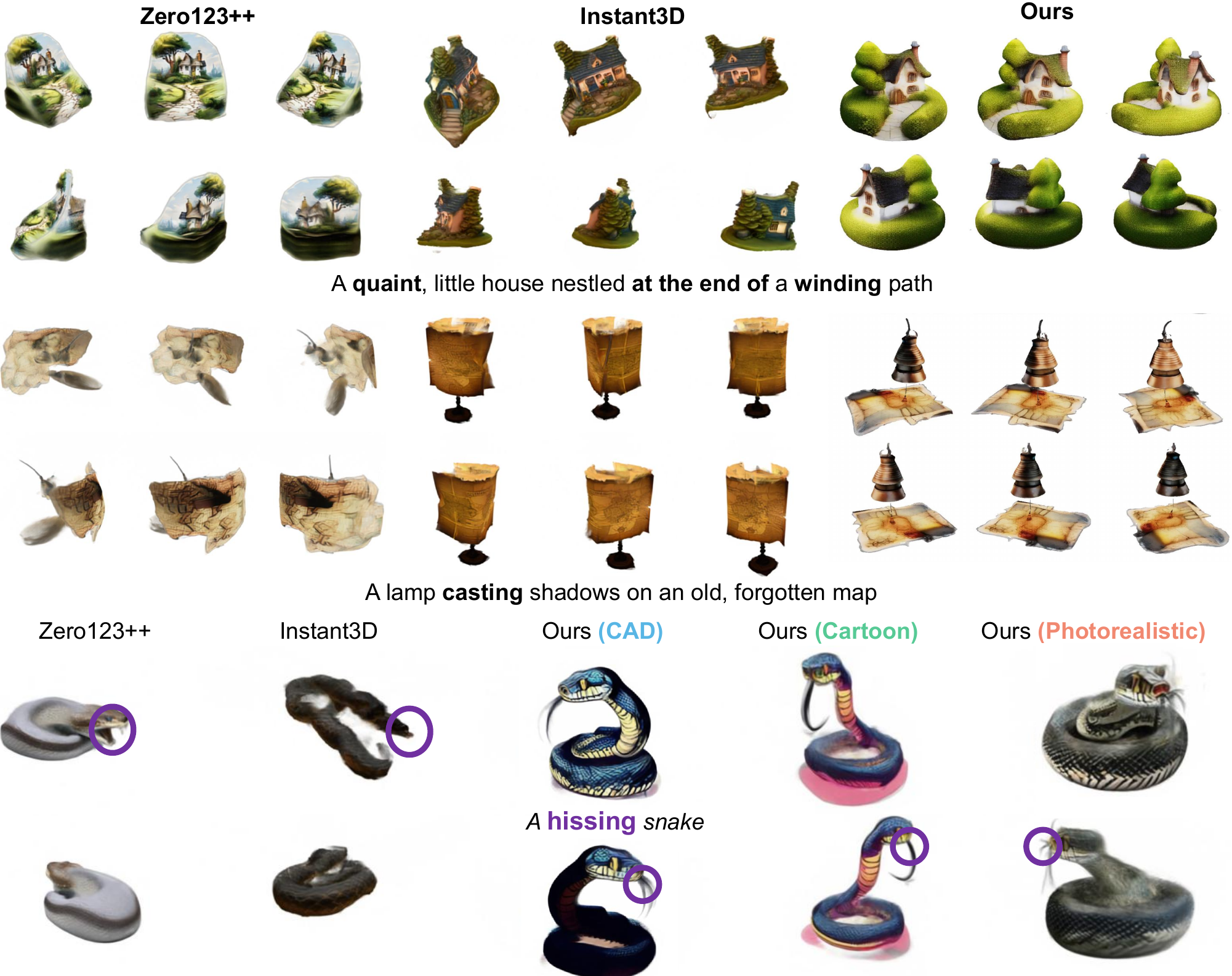}
  \vspace{-8mm}
\caption{\textbf{\methodname generates 3D objects compared to other edge-cutting methods} given text prompt. More results with higher resolution are available in Sup.\ref{sup_more_com}.}
   \vspace{-4mm}
   \label{fig:main_com}
\end{figure*}

\section{Experiments}

\subsection{Experiment Settings}
\textbf{Training data.}
For each set of 4-view images  obtained from both Objaverse~\citep{deitke2023objaverse} and generated by SV3D~\citep{voleti2024sv3d} or Zero123++\citep{shi2023zero123++}, we use MV-LLaVA to generate long descriptive captions with predicted quality score. Detailed quality check of MV-LLaVA is supplied in Sup.~\ref{sec:quality_of_mvllava} and data analysis in Sup.~\ref{sup_data_analysis}. In the end, we generate 200K 4-view image-text pairs on Objaverse~\citep{deitke2023objaverse}, 1000K 4-view image-text pairs from synthetic data from SV3D~\citep{voleti2024sv3d} and Zero123++\citep{shi2023zero123++}. We also sample 35K HQ SA~\citep{kirillov2023segment} data with captions from ShareGPT4V~\citep{chen2023sharegpt4v}.

\textbf{Training details.}
\label{sec:training_setting}
We test our framework directly on the text-to-multi-view diffusion model. We fine-tune PixArt-$\alpha$~\citep{chen2023pixart} with backbone DiT-XL/2~\citep{peebles2023scalable} model on the data as mentioned earlier. Similar to Instant3D~\citep{li2023instant3d}, we train the diffusion model directly on 4-view images naturally arranged in a 2$\times$2 grid. For 4 same view images from SA~\citep{kirillov2023segment}, we limit training time step $t\in [0, 50]$. We limit synthetic multi-view images $t\in [200, 1000]$. Regarding 3D object-rendered images, we do not limit $t$ but sample more frequently in the range $[50, 200]$ as a complement. We set the total batch size to 1024 with the learning rate set to 8e-5 for 20K steps. Training is conducted on 32 NVIDIA A100-80G GPUs for 20 hours with Flan-T5-XXL~\citep{chung2024scaling} text features and VAE~\citep{kingma2013auto} features pre-extracted.

\textbf{Evaluation metrics.}
\label{sec:test_setting}
We primarily benchmark the quantitative results of our approach and other methods from two main dimensions: 1). \textbf{Image-text alignment} measured by CLIP score and CLIP-R score indicating the prompt follow ability of text-to-multi-view (T2MV) diffusion model. 2). \textbf{Quality of generated images} measured by FID~\citep{heusel2017gans}. Given the trend of decoupling multi-view image generation and sparse view reconstruction, we conduct tests separately on multi-view images by T2MV and rerendered images from generated 3D objects. To test the robustness and diversity of \methodname beyond prompts generated by GPT, we also collect real user prompts from public website, the details and test results are available in Sup.~\ref{sec:real_user_prompts}.

\textbf{Evaluation details.}
For CLIP-R Score and CLIP Score, we test on 110 text prompts from GPTeval3D~\citep{wu2024gpt} using different CLIP models (i.e., CLIP-L/14~\citep{radford2021learning} and CLIP-bigG~\citep{openclip}) following the same setting of Instant3D~\citep{li2023instant3d}. Regarding the FID~\citep{heusel2017gans} test, as there is no golden standard for HQ 3D objects, we follow the similar evaluation idea of PlayGround2.5~\citep{li2024playground} (PG2.5) to use powerful T2I model generated images to form ground truth (GT) distribution. We use curated prompts to guide powerful PixArt and PG2.5 to generate high-quality CAD-style images with a single object in the pure background. Rembg~\citep{rembg} is adopted to create white background object-centric images. We use the method proposed in GPTeval3D~\citep{wu2024gpt} to generate 3K prompts. For both PG-2.5 and PixArt, we generate 10 images for each prompt with different seeds, resulting in 30K images to form the GT distribution of high-quality CAD-style objects. 

\textbf{Comparing methods.} In addition to Instant3D~\citep{li2023instant3d} and MVDream~\citep{shi2023mvdream} as direct text-to-multi-view (T2MV) methods, we also adopt edge-cutting single image to multi-view (I2MV) methods CRM~\citep{wang2024crm}, SV3D~\citep{voleti2024sv3d} and Zero123++\citep{shi2023zero123++}. For these methods, we condition the diffusion model on the single view image generated by PixArt (prompted to generate CAD-style single object-centric image). The result of the CLIP score is 3 times averaged with different seeds. For FID, we use 3 different seeds for each of the 3K prompts to generate 9K images to test the distance with GT high-quality images.

\subsection{Evaluation of Multi-view Images}
\label{sec:q_t2mv}

\begin{table}[t]
    \caption{\textbf{Benchmark of CLIP and FID score of text-to-multi-view (T2MV) models} on generated 4 view images, CLIP score tests on 110 text prompts from GPTeval3D~\citep{wu2024gpt} while FID is measured with the distribution of 30K object-centric images generated by SOTA T2I models. For text-to-image-to-multi-view(T2I2MV), we input I2MV models with single view images generated by Pixart-$\alpha$, which superior single view image quality is marked in \colorbox{mycolor_green}{green}.}
  \centering
  \scalebox{0.84}{
    \begin{tabular}{llccccccc}
    \toprule
    \multirow{2}{*}{\bf{Domain}} & \multirow{2}{*}{\bf{Method}} & \multicolumn{2}{c}{CLIP-R Score $\uparrow$} & \multicolumn{2}{c}{CLIP Score $\uparrow$} & \multicolumn{2}{c}{FID $\downarrow$} \\
     &    & CLIP-L/14 & CLIP-bigG & CLIP-L/14 & CLIP-bigG & PG2.5 & PixArt-$\alpha$ \\
    \midrule
    \rowcolor{mycolor_green} T2I & PixArt-$\alpha$ & \textcolor{gray!70}{96.1}  & \textcolor{gray!70}{94.7}  & \textcolor{gray!70}{25.9}  & \textcolor{gray!70}{41.5}  & \textcolor{gray!70}{20.7}  & \textcolor{gray!70}{5.4} \\
    \midrule
    \multirow{3}{*}{T2I2MV} & SV3D  & 78.8  & 81.3  & 24.7  & 37.3  & 55.7  & 54.2 \\
    & CRM & 77.5  & 85.1  & 24.9  & 38.9  & 59.0  & 52.2 \\
    & Zero123++ & 78.0  & 84.5  & 24.2  & 36.9  & 53.2  & 49.3 \\
    \midrule
    \multirow{3}{*}{T2MV} & Instant3D (unofficial) & 83.6  & 91.1  & 25.6  & 39.2  & 83.2  & 77.9 \\
    & MVDream & 84.8  & 89.3  & 25.5  & 38.4  & 60.2  & 59.2 \\
    \rowcolor{pearDark!20} & \methodname & \textbf{88.8} & \textbf{92.5} & \textbf{25.8} & \textbf{40.1} & \textbf{42.4} & \textbf{31.0} \\
    \bottomrule
    \end{tabular}%
    }
    \vspace{-2mm}
  \label{tab:t2mv_test}%
\end{table}%

\begin{figure*}[t]
  \centering
\includegraphics[width=1.0\linewidth]{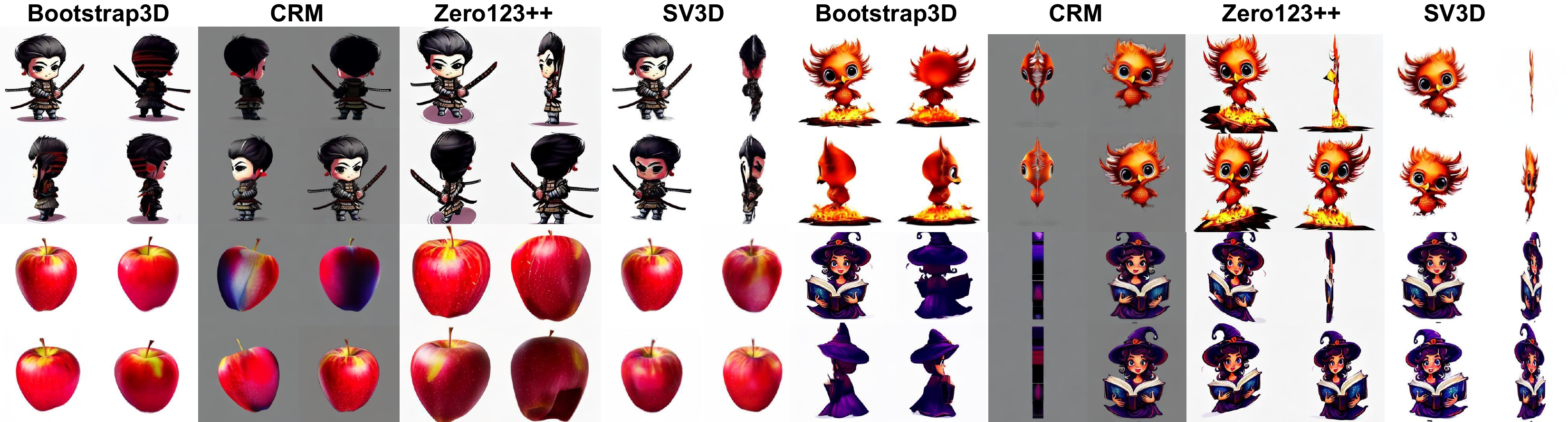}
  \vspace{-8mm}
\caption{\textbf{\methodname can generate high quality multi-view images} in out of domain cases compare to other edge-cutting multi-view diffusion models trained on Objaverse only.}
   \vspace{-6mm}
   \label{fig:mv_com}
\end{figure*}

As illustrated in Tab.\ref{tab:t2mv_test}, compared to other methods, the T2MV diffusion model trained by our framework yields the best results both according to image-text alignment and image quality. For qualitative experiments, we visualize some of the comparisons with other edge-cutting multi-view diffusion model in Fig.\ref{fig:mv_com}. For these image-to-multi-view models, we condition them on the top-left image generated by \methodname. 
Compared to these models trained solely on Objaverse~\cite{deitke2023objaverse}, our model demonstrates superior generalizability when the image domain is beyond the domain of Objaverse. Since it is difficult to directly measure view consistency as there is no ground truth 3D object for text-to-3D generation. we evaluate the view consistency by synthesizing 3D objects through large reconstruction model in the following experiments.
% It is worth noting that in the T2MV task, T2I2MV suffers a huge drop due to the domain gap between T2I-generated images and training images as well as the use of Rembg~\citep{rembg}. 
Qualitative results of real user cases are in Sup.~\ref{sec:real_user_prompts}.

\subsection{Evaluation of Generated 3D Objects}

% \begin{table}[htbp]
%   \centering
%   \caption{\textbf{Benchmark of CLIP and FID score of generated 3D objects} based on rendered 9 view images. *MVDream is tested on 200 generated objects for FID test using SDS~\citep{shi2023mvdream}, other methods are tested on 1000 objects using GRM~\citep{xu2024grm} as sparse view reconstruction model.}
%   \scalebox{0.82}{
%     \begin{tabular}{lcccccccc}
%     \toprule
%     \multirow{2}[4]{*}{\bf{Reconstruction}} & \multirow{2}[4]{*}{\bf{Method}} & \multicolumn{2}{c}{CLIP-R Score $\uparrow$} & \multicolumn{2}{c}{CLIP Score $\uparrow$} & \multicolumn{2}{c}{FID $\downarrow$} \\
% \cmidrule{3-8}        &  & CLIP-L/14 & CLIP-bigG & CLIP-L/14 & CLIP-bigG & PG2.5 & PixArt \\
%     \midrule
%     SDS & MVDream* & 85.2  & 90.8 & \textbf{26.1} & 39.4 & 57.8 & 56.7 \\
%     \midrule
%     \multirow{4}[4]{*}{GRM} & Instant3D (unofficial) & 81.7  & 89.4  & 24.8  & 37.1  & 85.4  & 80.3 \\
%     & SV3D & 74.1  & 82.8  & 23.4  & 34.1  & 68.4  & 69.1 \\
%     & Zero123++ & 71.2  & 80.3  & 22.3  & 34.5  & 69.3  & 72.4 \\
%     & \rowcolor{pearDark!20} \methodname & 86.3 & 91.6  & 25.9  & \textbf{39.7}  & \textbf{51.2}  & \textbf{50.7} \\
%     \midrule
%     \multirow{2}[2]{*}{InstantMesh} & Zero123++ & 73.2  & 84.1  & 23.0  & 37.2  & 82.3  & 88.8 \\
%     & \rowcolor{pearDark!20} \methodname & \textbf{87.1} & \textbf{92.0}  & 26.0  & 39.2  & 61.2  & 55.3 \\
%     \bottomrule
%     \end{tabular}%
%     }
%     \vspace{-4mm}
%   \label{tab:obj_test}%
% \end{table}%

\begin{table}[htbp]
  \centering
  \caption{\textbf{Benchmark of CLIP and FID score of generated 3D objects} based on rendered 9 view images. *MVDream is tested on 200 generated objects for FID test using SDS~\citep{shi2023mvdream}, other methods are tested on 1000 objects using GRM~\citep{xu2024grm} and InstantMesh~\citep{xu2024instantmesh} as sparse view reconstruction model.}
  \scalebox{0.82}{
    \begin{tabular}{lcccccccc}
    \toprule
    \multirow{2}[4]{*}{\bf{Reconstruction}} & \multirow{2}[4]{*}{\bf{Method}} & \multicolumn{2}{c}{CLIP-R Score $\uparrow$} & \multicolumn{2}{c}{CLIP Score $\uparrow$} & \multicolumn{2}{c}{FID $\downarrow$} \\
    \cmidrule(lr){3-8} &  & CLIP-L/14 & CLIP-bigG & CLIP-L/14 & CLIP-bigG & PG2.5 & PixArt \\
    \midrule
    SDS & MVDream* & 85.2  & 90.8 & \textbf{26.1} & 39.4 & 57.8 & 56.7 \\
    \midrule
    \multirow{4}[4]{*}{GRM} & Instant3D (unofficial) & 81.7  & 89.4  & 24.8  & 37.1  & 85.4  & 80.3 \\
    & SV3D & 74.1  & 82.8  & 23.4  & 34.1  & 68.4  & 69.1 \\
    & Zero123++ & 71.2  & 80.3  & 22.3  & 34.5  & 69.3  & 72.4 \\
    & \methodname & 86.3 & 91.6  & 25.9  & \textbf{39.7}  & \textbf{51.2}  & \textbf{50.7} \\
    \midrule
    \multirow{2}[2]{*}{InstantMesh} & Zero123++ & 73.2  & 84.1  & 23.0  & 37.2  & 82.3  & 88.8 \\
    & \methodname & \textbf{87.1} & \textbf{92.0}  & 26.0  & 39.2  & 61.2  & 55.3 \\
    \bottomrule
    \end{tabular}%
    }
    \vspace{-4mm}
  \label{tab:obj_test}%
\end{table}

% \begin{figure*}[h]
%   \centering
%   \includegraphics[width=1.0\linewidth]{figures/main_style_2.pdf}
%   \vspace{-6mm}
%    \caption{\textbf{\methodname can generate 3D objects with different style with more precise prompt control} compared to other edge-cutting methods. More results with thorough render views are available in Sup.\ref{sup_more_style}.}
%    \vspace{-6mm}
%    \label{fig:main_style}
% \end{figure*}

View consistency is another crucial factor in reconstructing reasonable 3D objects. Miss alignment between different views can lead to blurred areas in reconstructed objects by large reconstruction model~\citep{hong2023lrm,wei2024meshlrm}. This misalignment causes a significant deterioration in quality, resulting in a notable increase in metrics like FID. To assess the view consistency directly on 3D object, we employ GRM~\citep{xu2024grm} and InstantMesh~\citep{xu2024instantmesh} to reconstruct the object given sparse view images generated in Sec.~\ref{sec:q_t2mv}. We render 9 view images evenly in orbit for each object and evaluate the image-text alignment and image quality. As reported in Tab.~\ref{tab:obj_test}. \methodname, after conditioning GRM or InstantMesh on 4 view images, can generate the best 3D objects both according to image-text alignment and image quality. GPT-4V based human-aligned evaluation based on GPTeval3D~\citep{wu2024gpt} is supplied in Sup.~\ref{sup:gpt_evaluation}.

We also present visualizations of some results in Fig.\ref{fig:main_com}. \methodname can generate objects with higher quality and prompt following ability. For other methods, as shown in the first column of Fig.\ref{fig:main_com}, although the first image may be well aligned with the given text prompt, the final 3D object may be compromised due to the limitations of its poor generalizability as they are also fine-tuned on Objaverse~\citep{deitke2023objaverse} only. 
% The domain gap between the rendered Objaverse image and images generated by the T2I model limits the outcome of the text-to-single image, single-view-to-3D pipeline. 
More visualizations and discussions of this are in Sup.~\ref{sup:vis_compare}

\subsection{Ablation Study}

\begin{figure*}[h]
  \centering
  \includegraphics[width=1.0\linewidth]{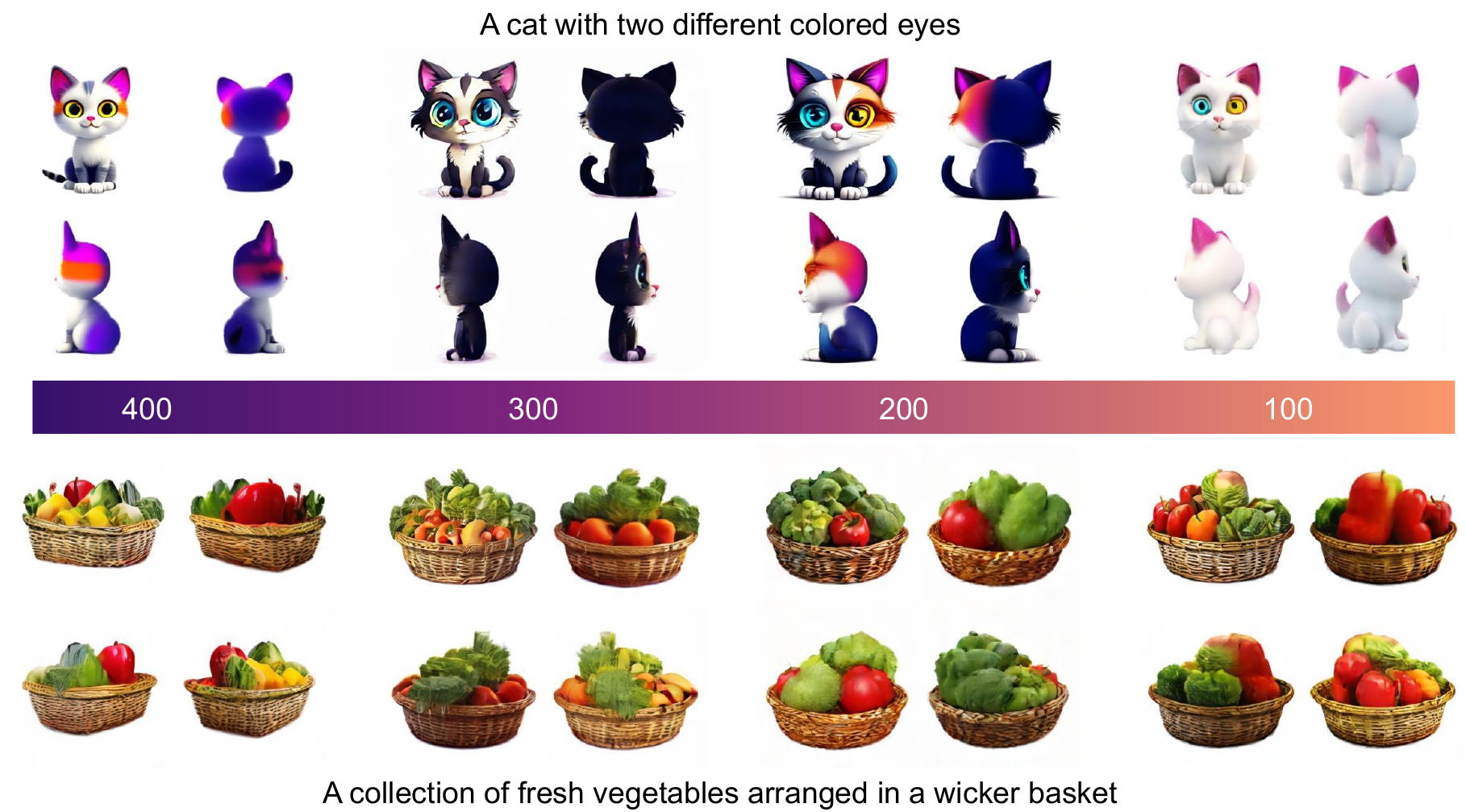}
  % \vspace{-10pt}
  \vspace{-8mm}
   \caption{\textbf{Ablation study of training time reschedule (TTR)} demonstrates a trade-off between image-text alignment and image quality with different $t$.}
   \vspace{-6mm}
   \label{fig:sup_more_com}
\end{figure*}

\textbf{Training Timestep Reschedule (TTR)} is proposed in \ref{method_tnr} to better integrate different types of data.  The training time step of synthetic data is restricted in $[T, 1000]$, where $T$ is a hyper-parameter to be set in training. We demonstrate the effect of the time-step limit in Fig.\ref{fig:sup_more_com}, where the bar in the middle is the value of $T$. When $T$ is large, namely synthetic data won't affect more time-step at the end of the denoising process, Synthetic data has less influence on the denoising process towards the end, which leads to better view consistency but lower prompt-following ability. Conversely, if $T$ is small, the denoised result better follows the given text prompt but blurring becomes much more severe. In summary, there is a trade-off in injecting synthetic data into the training process: better image-text alignment comes at the cost of worse view consistency and increased blurring. Ultimately, we set $T=200$ based on empirical study.

\begin{table}[htbp]
  \centering
  \caption{\textbf{Ablation study of proposed components and quantity of synthetic data.} with CLIP-R Score represents image-text alignment and FID represents image quality.}
  \scalebox{0.88}{
    \begin{tabular}{lcccc}
    \toprule
    \multicolumn{1}{c}{\multirow{2}[0]{*}{Methods}} & \multicolumn{2}{c}{Multi-view Image} & \multicolumn{2}{c}{Generated Object} \\
          & CLIP-R Score & FID PG-2.5 & CLIP-R Score & FID PG-2.5 \\
    \midrule
    Instant3D (unofficial)& 83.6  & 83.2  & 81.7  & 85.4 \\
    \midrule
    Cap3D only & 77.9  & 101.3 & 74.6  & 120.4 \\
    Cap3D + Synthetic Image (100k) w/o TTR & 81.5  & 92.0  & 71.2  & 134.6 \\
    Cap3D + Synthetic Image (100k) w/ TTR & 83.3  & 60.8  & 80.2  & 70.6 \\
    Dense recaption + Synthetic Image (100k) & 87.4  & 50.2  & 85.1  & 50.9 \\
    Dense recaption + Synthetic Image (500k) & 88.8  & 42.4  & 86.3  & 51.2 \\
    \bottomrule
    \end{tabular}%
    }
  \vspace{-3mm}
  \label{tab:ablation_component}%
  
\end{table}%

\textbf{Synthetic data and dense captioning} are proposed in our work to achieve high-quality images and better image-text alignment. We ablate their effects and the importance of data quantity in Tab.~\ref{tab:ablation_component}. Direct use of synthetic data without Training Timestep Reschedule (TTR) can cause severe blurs and deformation in final outcome. With the help of TTR, the mixture of data can not only improve image-text alignment but also maintain view consistency. Replacing Cap3D~\citep{luo2024scalable}'s caption with MV-LLaVA's dense descriptive caption further improves the model's capability of following prompts faithfully. Improvement through increasing volume of data also proves the scalability of our framework.

\section{Conclusion and Discussion}
In this work, we introduce a novel framework that employs MLLMs and diffusion models to synthesize high-quality data for bootstrapping multi-view diffusion models. With a powerful fine-tuned 3D-aware MLLM serving as the dense captioner and quality filter, the generated synthetic data addresses the issue of insufficient high-quality 3D data. The proposed strategy of injecting different data at different training time steps uses the property of the denoising process to further achieve higher image quality while maintaining view consistency. We believe this work will contribute to the goal of achieving 3D content creation with each rendered view comparable with the single view diffusion model, with more advanced MLLMs and diffusion models on the horizon.

\label{sec:limitations}
\textbf{Limitations and future work.} Despite its promise, our work still faces several unresolved challenges. Firstly, the multi-view diffusion model is only the first step of the 3D content creation pipeline. Sparse view reconstruction models also need improvement as most edge-cutting sparse view reconstruction models are also trained on Objaverse~\cite{deitke2023objaverse} only. Secondly, Although MLLMs can estimate general quality and view consistency, subtle view inconsistency is hard to detect until ambiguity leads to blurred areas in reconstructed 3D object. While the proposed Training Timestep Reschedule can mitigate this problem, it cannot solve the problem fundamentally. Using synthetic data to train sparse view reconstruction models and quality estimation directly based on the reconstructed object are thus interesting future directions for improving 3D content creation.

\bibliography{iclr2025_conference}
\bibliographystyle{iclr2025_conference}

\newpage
\appendix
\section{Appendix}

\subsection{Evaluation on wild prompts from real users}

\label{sec:real_user_prompts}
\begin{figure*}[h]
  \centering
  \includegraphics[width=1.0\linewidth]{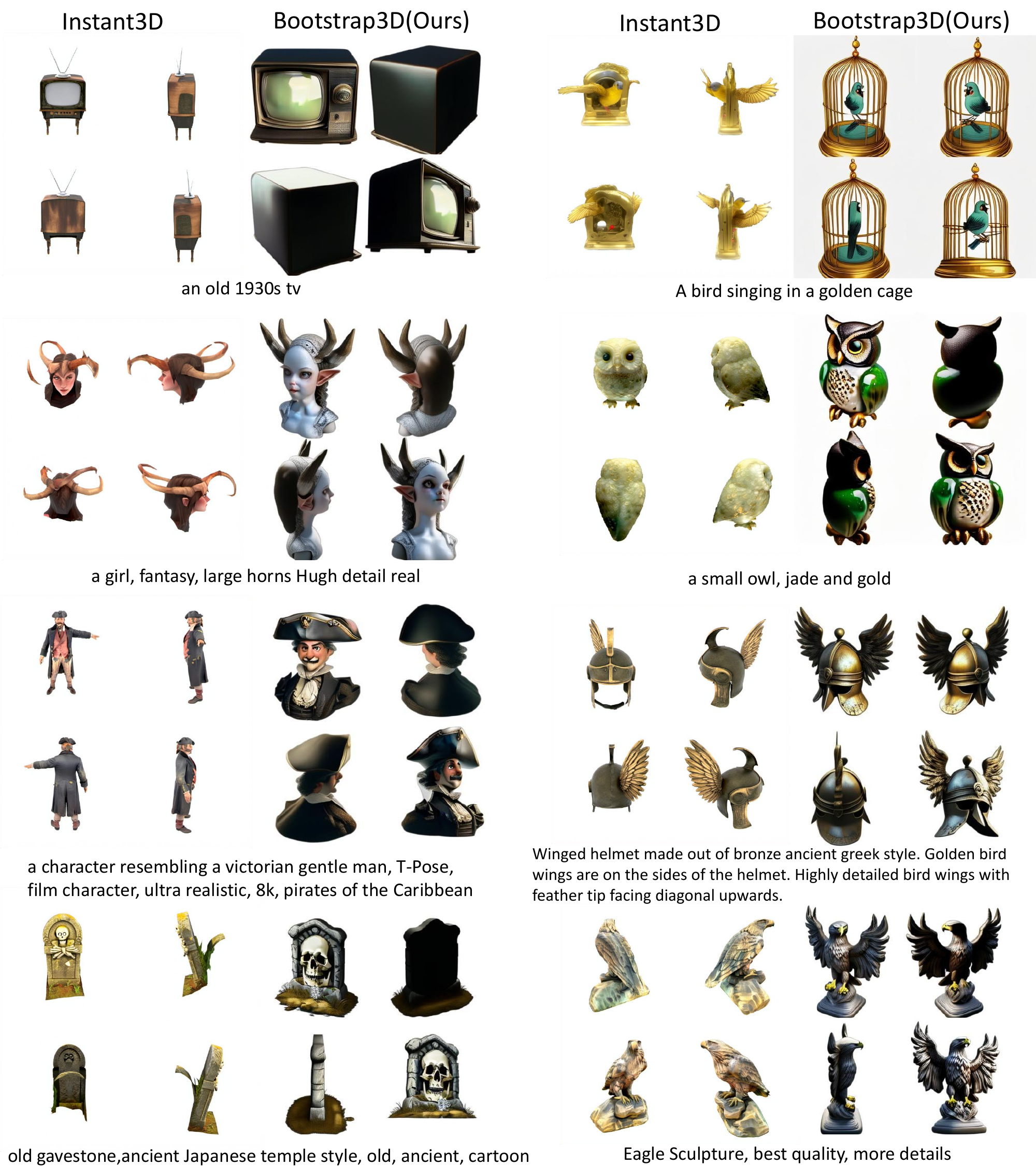}
   \caption{\textbf{Real user prompt cases} visualization compared to Instant3D~\citep{li2023instant3d}} 
   \label{fig:wild_cases}
   \vspace{-4mm}
\end{figure*}

The results of the main part of the paper are only tested on GPT generated prompts. To test our work's capability in wild cases, we also collect real user prompts and compare our method with Instant3D~\citep{li2023instant3d}. specifically, we randomly collect 100 prompts from \url{https://www.meshy.ai/} and test the CLIP-R precision as well as GPT based evaluation (detailed in Sup.~\ref{sup:gpt_evaluation}). Results and some qualitative cases are shown in Tab.~\ref{tab:wild_cases} and Fig.~\ref{fig:wild_cases}. We highlight that our \methodname excels Instant3D~\citep{li2023instant3d}
when tested on real user prompts through training on synthetic data.

\begin{table}[!ht]
    \centering
    \vspace{-4mm}
    \caption{\textbf{Test results of in the wild cases.} \methodname also excels Instant3D~\citep{li2023instant3d} in generating high quality images according to real user prompts.} 
    \begin{tabular}{lccc}
    \toprule
        \multirow{2}{*}{\textbf{Method}} & CLIP based metric & \multicolumn{2}{c}{GPTEval3D}  \\ 
        ~ & CLIP-R score & image-text alignment & texture detail \\ 
        \midrule
        Instant3D (unofficial)  & 77.0 & 22.0\% & 24.5\% \\ 
        Bootstrap3D & 83.5 & 78.0\% & 75.5\% \\ 
        \bottomrule
    \end{tabular}
    \vspace{-4mm}
    \label{tab:wild_cases}
\end{table}

\begin{figure*}[h]
  \centering
\includegraphics[width=1.0\linewidth]{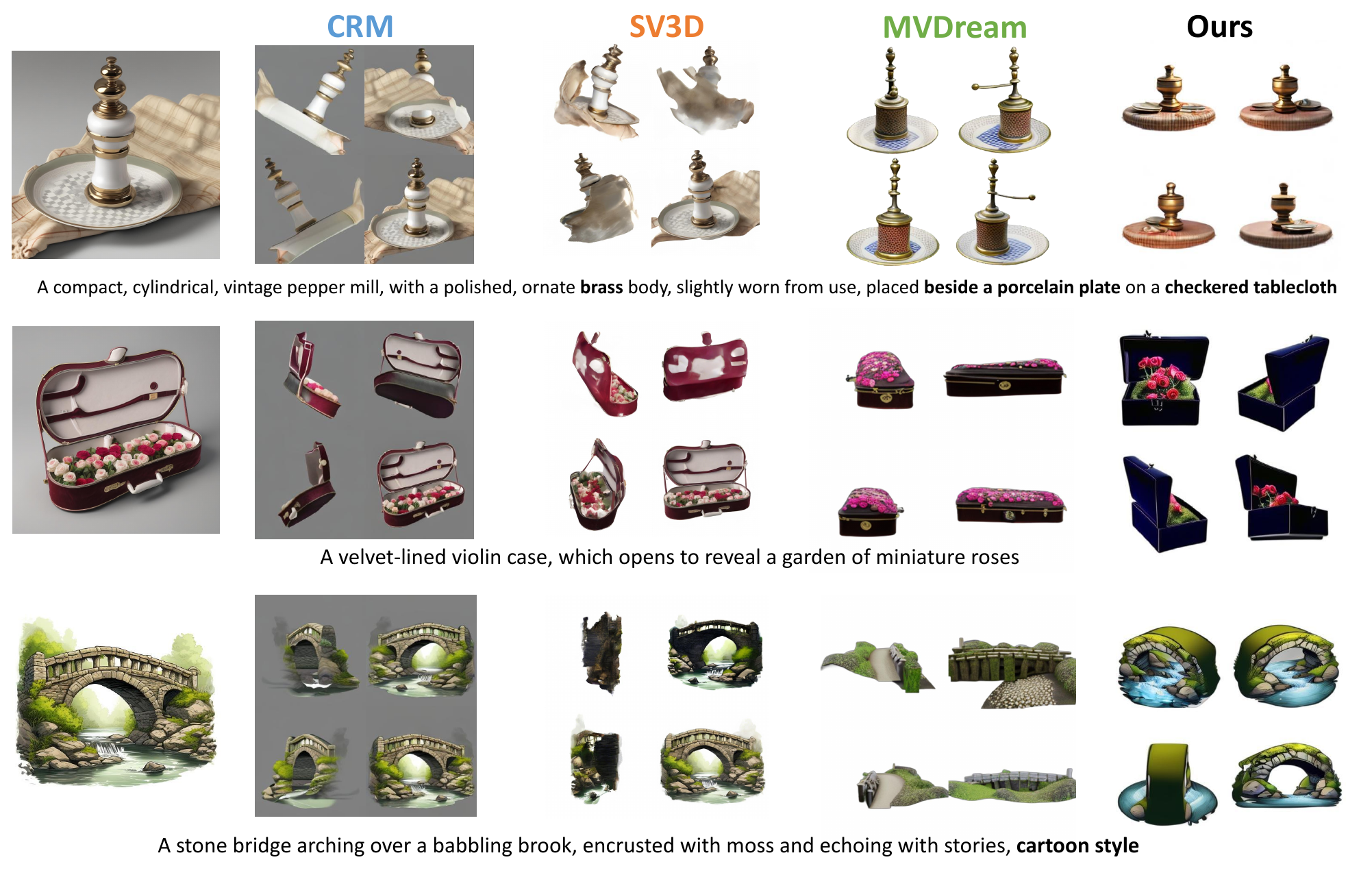}
   \caption{\textbf{Generated multivew images compare to other methods.} Our method can generate multi-view images with long text control without encountering blurring effect from data generated by SV3D thanks to TTR and quality filtering.}
   \label{fig:more_visual}
   \vspace{-4mm}
\end{figure*}

\subsection{More Visualization Compared to other Methods.}
\label{sup:vis_compare}
We show more visualization of the quantitative experiments shown in the main paper in Fig.\ref{fig:more_visual}

For Image-to-3D methods, they can sometimes produces significant motion blurring and fails when the input image is out-of-distribution (like the 3rd cartoon style case). We resample the high-quality segment of the distribution of generated images using quality filtering based on MLLM methods. Furthermore, by employing TTR, we limit the impact of these data when training multi-view diffusion models, allowing our model to produce much clear results. In addition, we use a caption rewriting method, enabling finer prompt control for the generated multi-view images.

\subsection{Data Statistics}
\label{sup_data_analysis}
\subsubsection{Caption Analysis}

\begin{figure*}[h]
  \centering
  \includegraphics[width=0.86\linewidth]{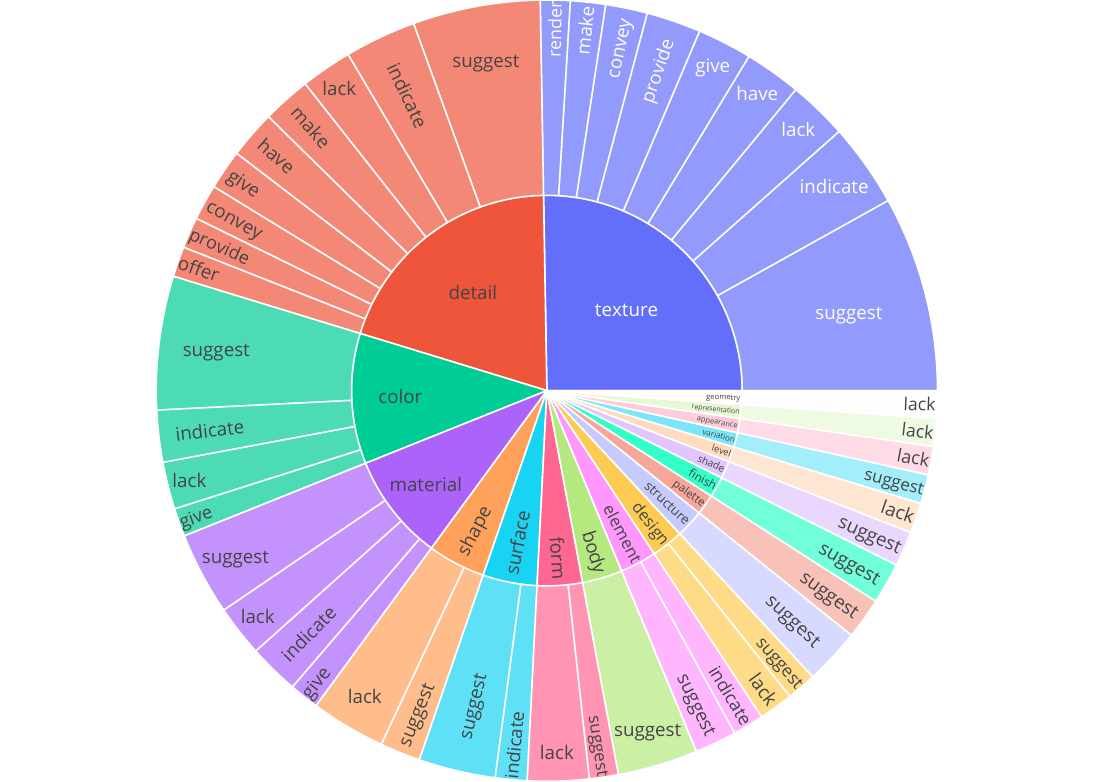}
   \caption{\textbf{Visualized analysis of dense reasoning descriptions generated by GPT4-Vision~\citep{2023GPT4VisionSC}} of the root noun-verb pairs (occurring over 1\%) of the descriptions} 
   \label{fig:prompt_gpt}
\end{figure*}

\begin{figure*}[h]
  \centering
\includegraphics[width=0.86\linewidth]{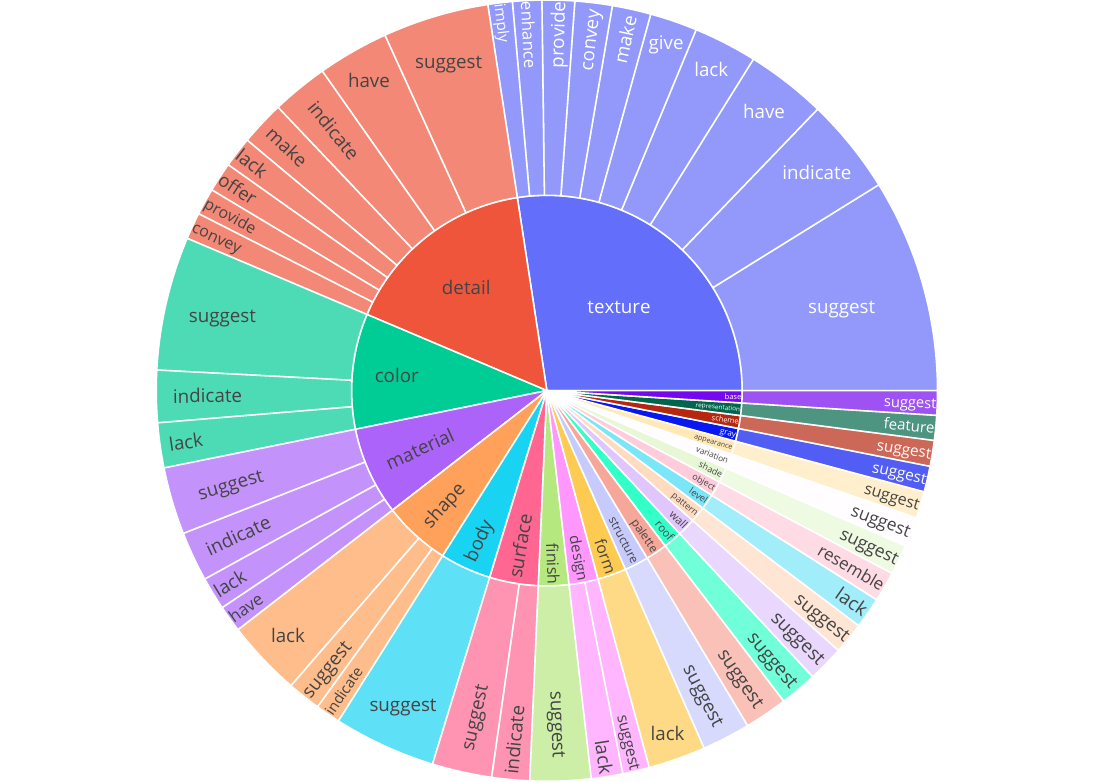}
   \caption{\textbf{Visualized analysis of dense reasoning descriptions generated by our MV-LLaVA} of the root noun-verb pairs (occurring over 1\%) of the descriptions} 
   \label{fig:prompt_llava}
\end{figure*}

Fig.~\ref{fig:prompt_gpt} and ~\ref{fig:prompt_llava} provide a visualization of the root noun-verb pairs for the captions generated by GPT-4V~\citep{2023GPT4VisionSC} and MV-LLaVA.  It’s clear to see that the diversity and linguistic
expression of the captions produced by MV-LLaVA are
highly matched with those of GPT-4V. We believe the highly detailed description focusing on object's texture, shape and color have potential usage beyond training multi-view diffusion model in the field like object texturing~\citep{fang2024makeitreal} and stylization~\citep{sharma2023alchemist} in Computer Graphics. MV-LLaVA can also serve as free and efficient 3D object assistant comparable with GPT-4V for future research of 3D content creation.

\begin{figure*}[h]
  \centering
  \includegraphics[width=1.0\linewidth]{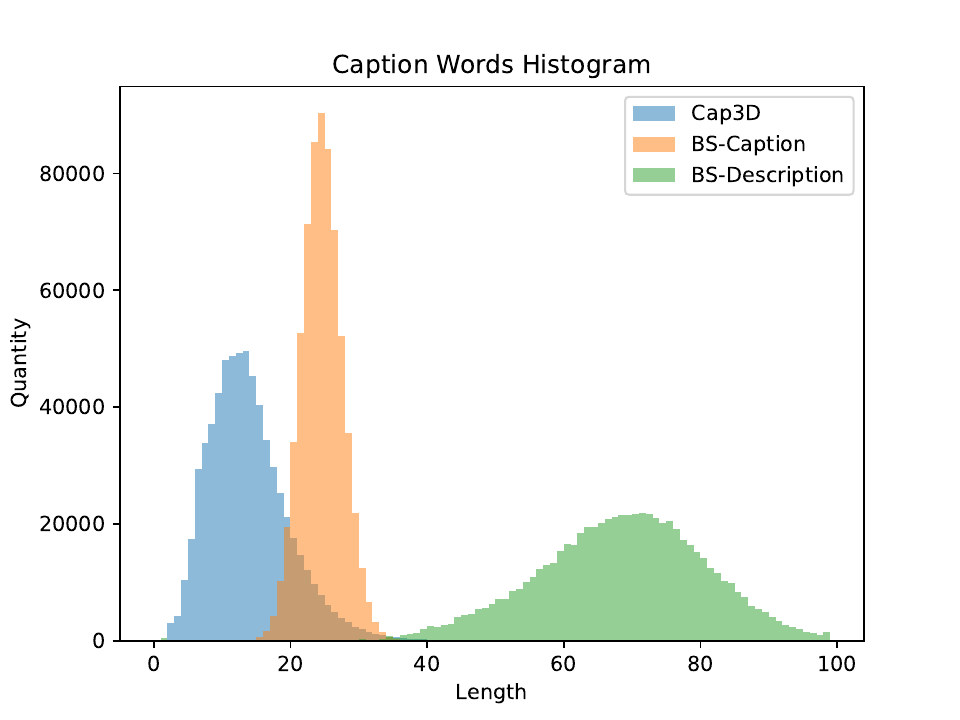}
   \caption{\textbf{Histogram Visualization of the Caption Length} compared with Cap3D~\citep{luo2024scalable}}
   \label{fig:cap_hist}
\end{figure*}

\begin{table}[h]
\centering
\caption{\textbf{Comparison of lexical composition of the captions} generated by GPT4-Vision and Share-Captioner.} 
% \resizebox{1\linewidth}{!}{
% \setlength{\tabcolsep}{1mm}{
\begin{tabular}{l|cccccc}
\toprule
Lexical &n. &adj. &adv. &v. &num. &prep.\\ \midrule
GPT-4V~\citep{2023GPT4VisionSC} &29.1$\%$ &16.0$\%$ &1.5$\%$ &11.1$\%$ &0.5$\%$ &9.0$\%$ \\ \midrule
BS-Description &28.5$\%$ &16.0$\%$ &1.4$\%$ &10.8$\%$ &0.3$\%$ &8.6$\%$ \\ 
BS-Caption &30.2$\%$ &23.0$\%$ &0.3$\%$ &5.6$\%$ &0.1$\%$ &8.9$\%$ \\ \bottomrule
\end{tabular}
% }
% }
% \vspace{-2mm}
\label{table:lexical}
\end{table}

Fig.~\ref{fig:cap_hist} visualizes the histogram of caption length compared with Cap3D~\citep{luo2024scalable}. We fine-tune MV-LLaVA to generate two different lengths suitable for different diffusion architecture, namely CLIP-based text encoding~\citep{blattmann2023stable,podell2023sdxl} with 77 token length and T5 based text encoding~\citep{chen2023pixart,chen2024pixart} with 120 token length. Both excel the length of Cap3D with less hallucinations.

\subsubsection{Estimated Quality Analysis}
\begin{figure*}[h]
  \centering
  \includegraphics[width=0.99\linewidth]{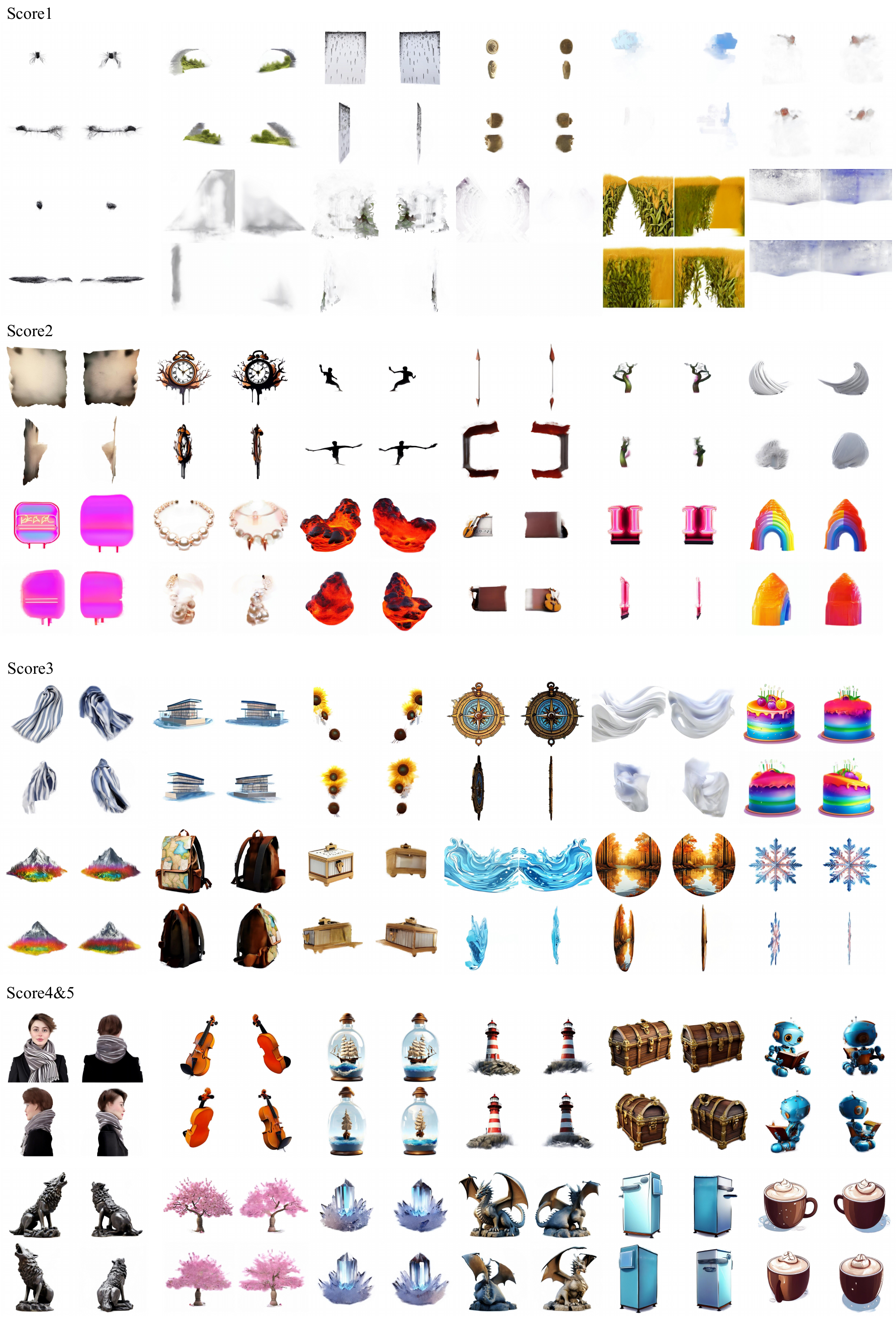}
  \vspace{-5pt}
   \caption{\textbf{Randomly picked multi-view images with different scores from 500k synthetic data generated by SV3D~\citep{voleti2024sv3d}.}}
   \label{fig:quality_sv3d}
\end{figure*}

\begin{figure*}[h]
  \centering
  \includegraphics[width=0.99\linewidth]{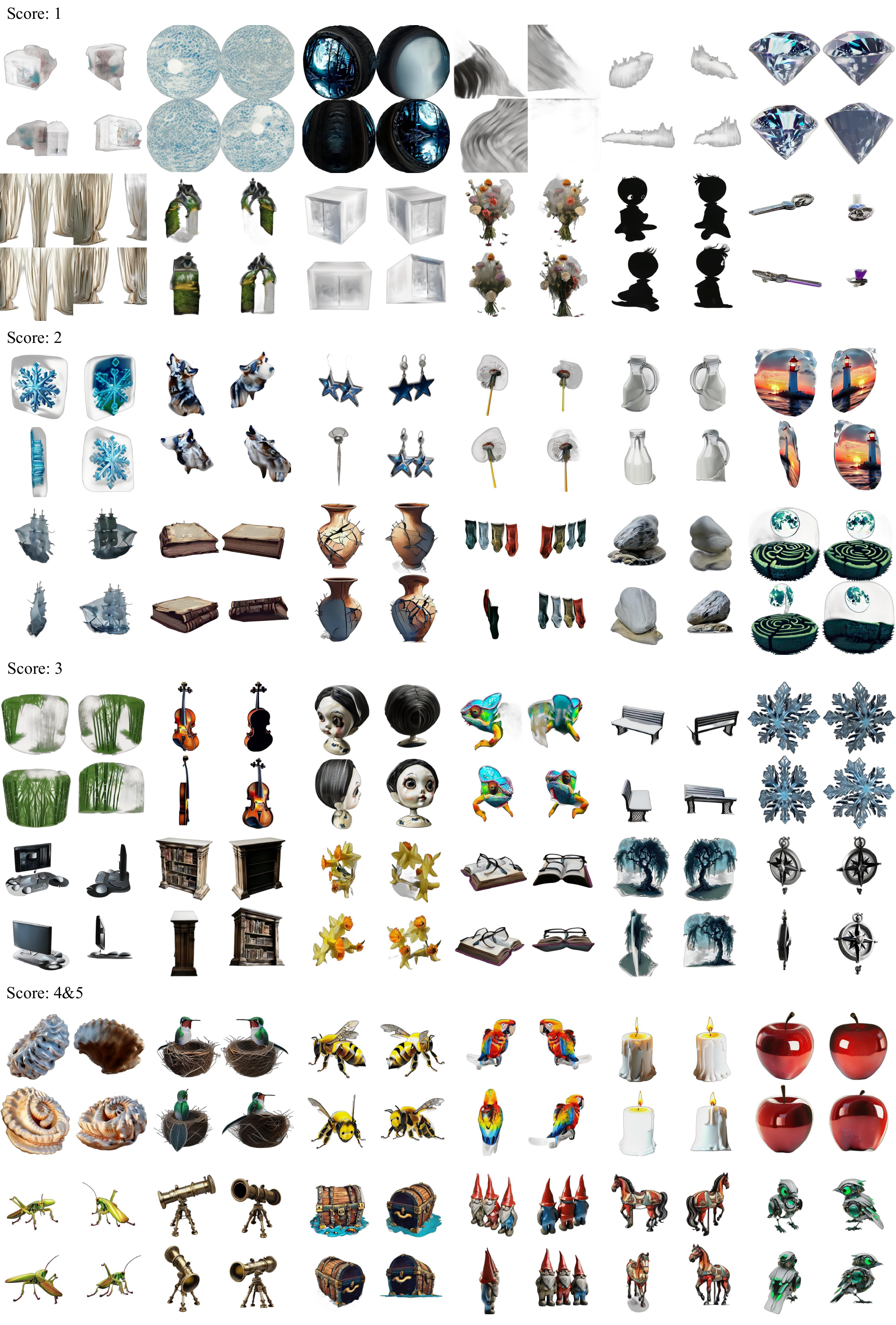}
  \vspace{-5pt}
   \caption{\textbf{Randomly picked multi-view images with different scores from 500k synthetic data generated by Zero123++~\citep{shi2023zero123++}.}}
   \label{fig:quality_zeropp}
\end{figure*}

\begin{figure*}[h]
  \centering
  \includegraphics[width=0.99\linewidth]{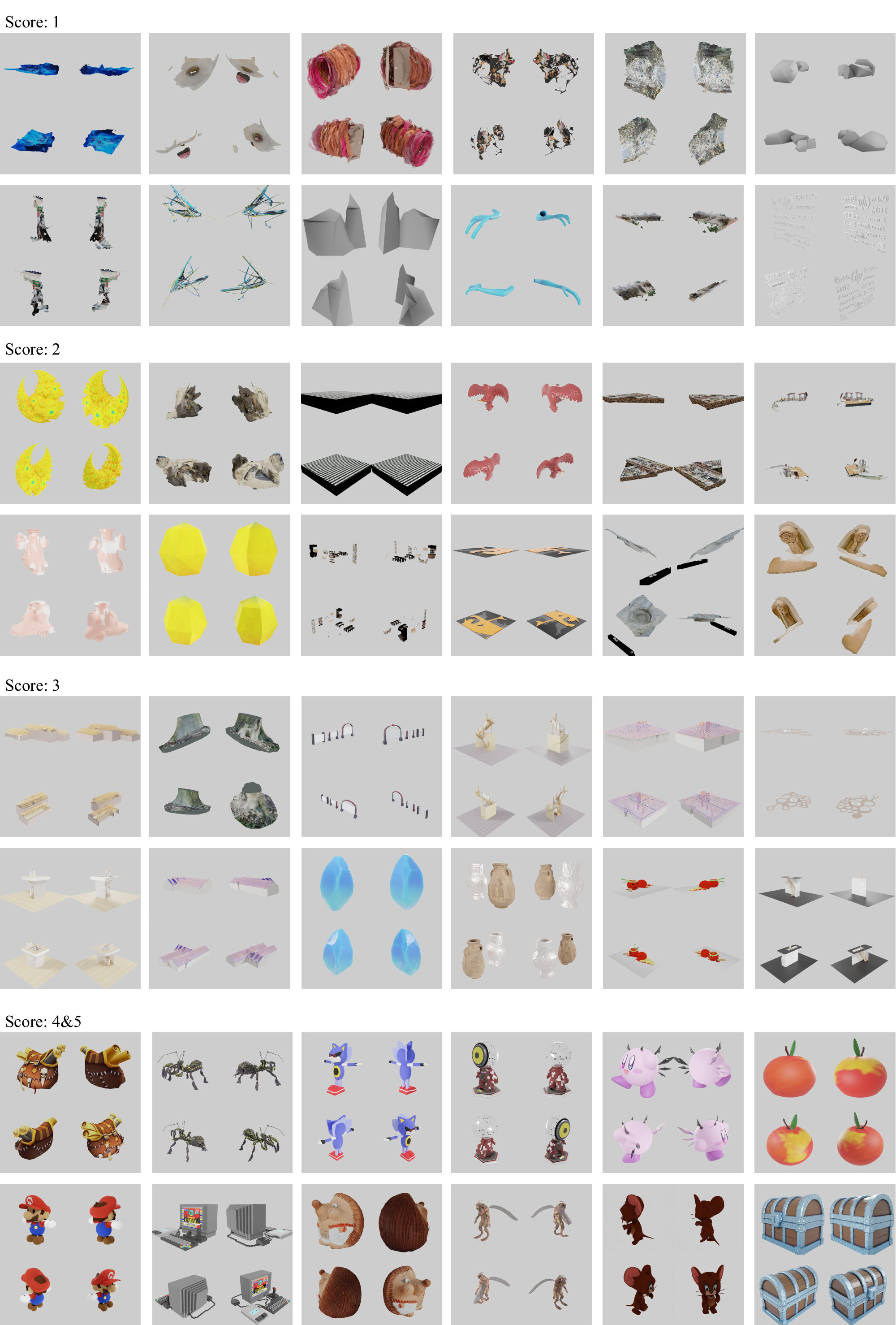}
   \caption{\textbf{Randomly picked multi-view images with different scores from 660k Objaverse~\citep{deitke2023objaverse} 3D assets.}}
   \label{fig:quality_cap3d}
\end{figure*}

For direct grasp of the quality of objaverse data and synthetic data used to train diffusion model, we randomly picked some of multi-view images from different score rank. Results are shown in Fig.~\ref{fig:quality_sv3d}, Fig.~\ref{fig:quality_zeropp} and Fig.~\ref{fig:quality_cap3d}. We use high quality data with score 4 and 5 for the training of multi-view diffusion model.

\begin{figure*}[h]
  \centering
  \includegraphics[width=1.0\linewidth]{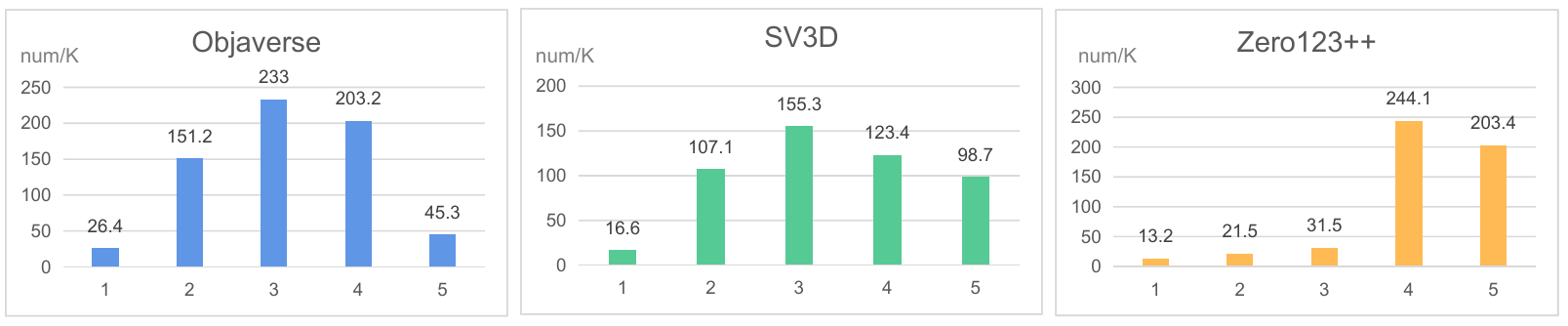}
   \caption{\textbf{Quality score statistics of different data source.}}
   \label{fig:quality_statistic}
\end{figure*}
We count the number of multi-view images from different data sources, namely 660K from Objaverse, 500K from SV3D~\citep{voleti2024sv3d} and 500K from Zero123++~\citep{shi2023zero123++} generated by our \methodname pipeline. Result are shown in Fig.\ref{fig:quality_statistic}. For Objaverse and SV3D, the assigned score is normal and we use score 4 and score 5 multi-view images as high quality data for training. However, for Zero123++, most objects are assigned with score greater than 3. We attribute this phenomenon to the fact that Zero123++ tend to generate objects with less motion blurring but more stretching and deformation compared to SV3D. Joint training of MV-LLaVA on three different data source lead to higher and more focused distribution for Zero123++'s multi-view images. For this part of synthetic data, we leave only score 5 multi-view images as high quality data.

\subsection{Quality of MV-LLaVA}
\label{sec:quality_of_mvllava}
\subsubsection{Choice of number of unfrozen layers of vision encoder.}
\label{sup:mvllava_layers}
\begin{figure*}[h]
  \centering
  \includegraphics[width=0.99\linewidth]{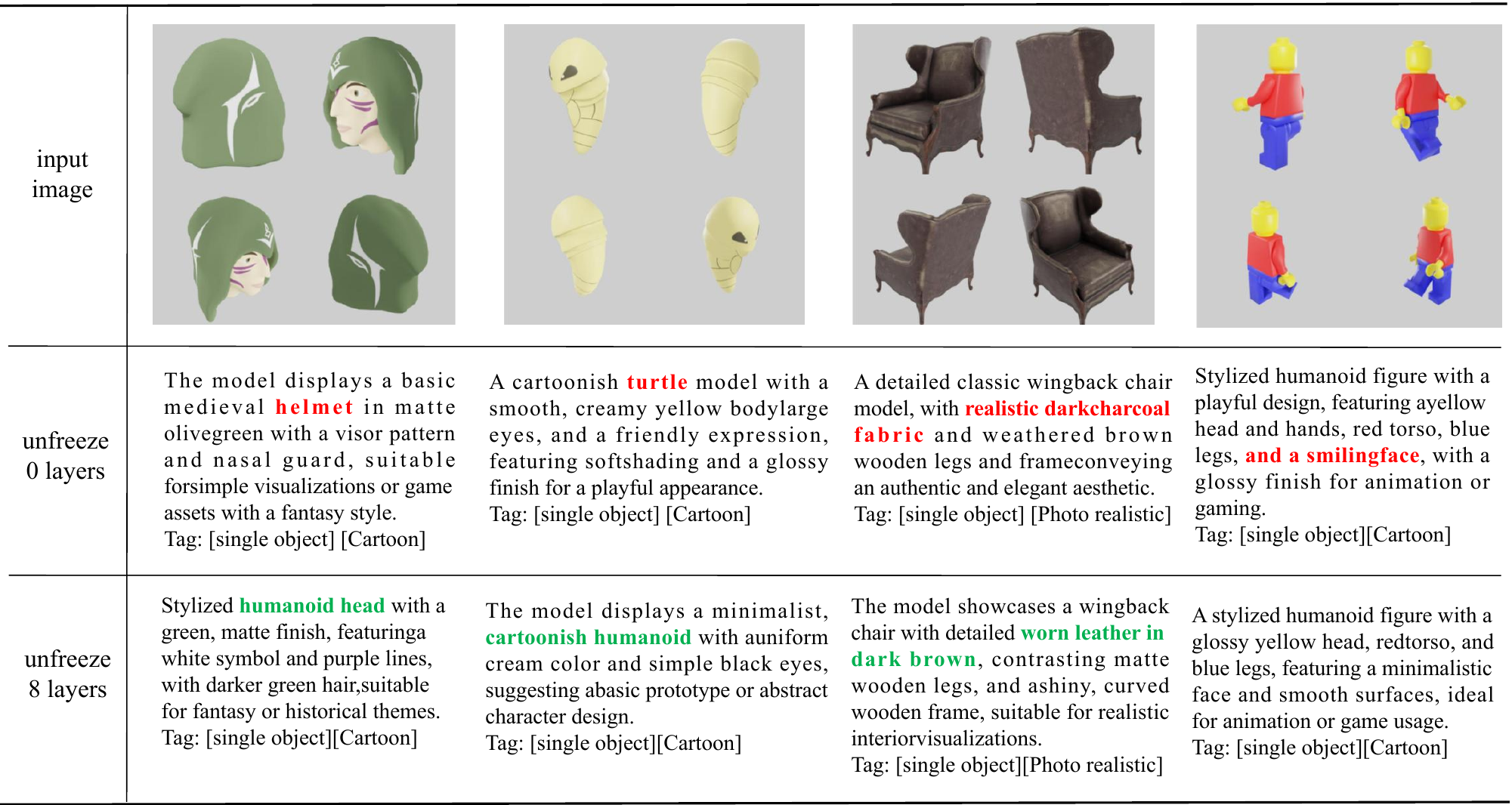}
  \vspace{-5pt}
   \caption{\textbf{Qualitative results of unfreeze final layers of CLIP~\citep{radford2021learning} vision encoder} compared to original fixed vision encoder setting in LLaVA~\citep{liu2024visual}.}
   \label{fig:unfreeze_layers}
\end{figure*}

Inspired by ShareGPT-4V~\citep{chen2023sharegpt4v}, we unfreeze selected final layers of the CLIP~\citep{liu2024visual} vision encoder during the initial phase of vision language alignment. The CLIP-L/14 model used for LLaVA~\citep{liu2024visual} contains 24 transformer layers. We selectively unfreeze some of final layers to enable the CLIP model to focus more on details such as texture of multi-view images. After qualitative manual screening, we select to unfreeze eight layers to yield better results. Fig.~\ref{fig:unfreeze_layers} illustrates the differences between unfreezing eight layers and not unfreezing any (the original training setting of LLaVA~\citep{liu2024visual}). The red sections highlight the erroneous hallucinations occurring when the vision encoder remains fully unchanged, while the green sections indicate accurate descriptions of the image content. This demonstrates that partially unfreezing the vision encoder can produce more precise captions and reduce some hallucinations.

\subsubsection{Quantitative quality study}
To test the quality of our MV-LLaVA. We propose two quantitative study over the quality of captions and the alignment of quality estimation with human experts. In first study, we randomly picked 200 object from Objaverse~\citep{deitke2023objaverse} and exclude training data of MV-LLaVA. We use GPT4-V~\citep{2023GPT4VisionSC} and MV-LLaVA to generate descriptive captions for each object. We invite human volunteers to choose their preference over shuffled captions. Results are shown in Tab.~\ref{table:user_study}, where MV-LLaVA shows comparable captioning ability with powerful GPT4-V~\citep{2023GPT4VisionSC}, which is essential to generate millions of high quality image-text pairs for the training of text to multi-view image diffusion model.

Second experiment studies MV-LLaVA's ability in quality estimation of both 3D assets and generated multi-view images. We invite human volunteers to estimate the quality of multi-view images rendered from Objaverse~\citep{deitke2023objaverse} or generated by SV3D~\citep{voleti2024sv3d}. As there is no golden standard for multi quality classification, We ask them to separate the randomly select multi-view images into approximately two half and serve as GT quality. We use MV-LLaVA to estimate the quality of these images and generate confusion matrix. Results are shown in Tab.\ref{tab:confusion_matrix_mv}. Given the great amount of source data of 3D assets and infinite synthetic data, we care more about the false positive rate, as these data will be mixed into training data. In this observation, we highlight the false positive rate of over 20\% for SV3D~\citep{voleti2024sv3d} generated multi-view images. This result align with the observation of inevitable motion blurring of SV3D~\citep{voleti2024sv3d}. To leverage this part of data source for data diversity without hurting the final quality. We propose Training Noise Reschedule to avoid samplings from these synthetic data when time step is small.

\begin{table}[h]
\centering
\caption{\textbf{Human evaluation} on the quality of generated captions from MV-LLaVA vs. GPT4-Vision~\citep{2023GPT4VisionSC} over 200 validation samples from Objaverse~\citep{deitke2023objaverse}.} 
\begin{tabular}{l|ccc}
\toprule
Preference & GPT4-Vision~\citep{2023GPT4VisionSC} & MV-LLaVA & Comparable \\ \midrule
Percentage &39.5$\%$  &34.5$\%$  &26.0$\%$  \\ \bottomrule
\end{tabular}
\label{table:user_study}
\end{table}

\begin{table}[htbp]
  \centering
  \caption{\textbf{Confusion matrix} of mutli-view images quality estimation.}
    \begin{tabular}{c|cc|c|cc}
    \toprule
    \multicolumn{3}{c}{Objaverse quality check} & \multicolumn{3}{c}{Synthetic quality check} \\
    \midrule
          & HQ-gt & LQ-gt &       & HQ-gt & LQ-gt \\
    \midrule
    HQ by model & 31.0\% & 4.5\% & HQ by model & 34.5\% & 11.5\% \\
    LQ by model & 11.0\% & 53.5\% & LQ by model & 17.0\% & 37.0\% \\
    \bottomrule
    \end{tabular}%
  \label{tab:confusion_matrix_mv}%
\end{table}%

\subsubsection{Qualitative caption quality study}
\begin{figure*}[h]
  \centering
  \includegraphics[width=0.99\linewidth]{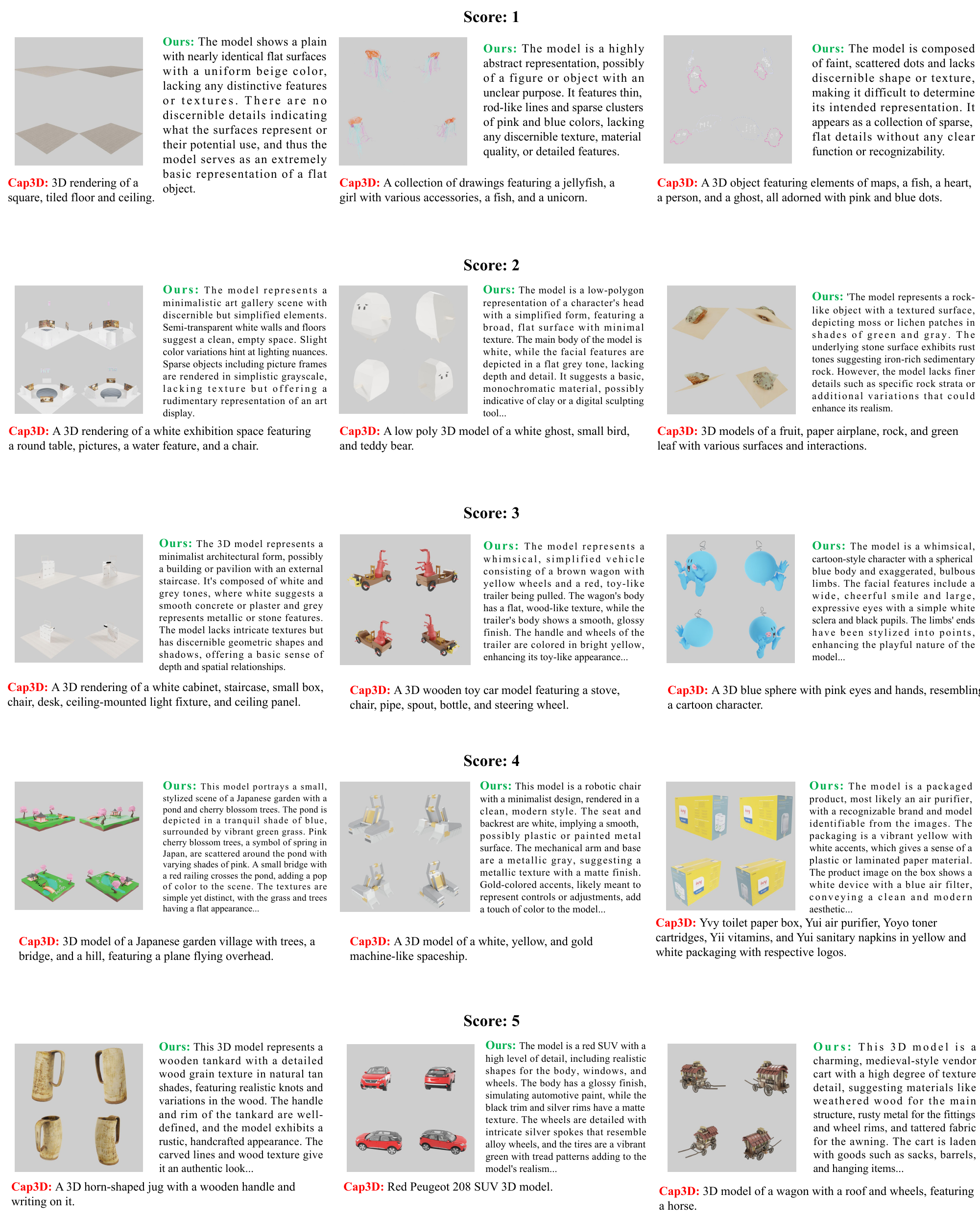}
   \caption{\textbf{Caption comparison with Cap3D~\citep{luo2024scalable}.} Our MV-LLaVA can generate long captions that faithfully describing 3D assets from different perspectives like color, geometry and texture.}
   \label{fig:caption_compare}
\end{figure*}

We selective compare some of the captions generated by Cap3D~\citep{luo2024scalable} and MV-LLaVA in Fig.~\ref{fig:caption_compare}. Our MV-LLaVA can generate more detailed descriptive captions with less hallucinations.

\clearpage
\newpage

\subsection{Details of Prompt Design}

\subsubsection{Prompts for GPT-4V for Quality Check}
\label{prompt_detail}
\begin{figure*}[h]
  \centering
  \includegraphics[width=0.99\linewidth]{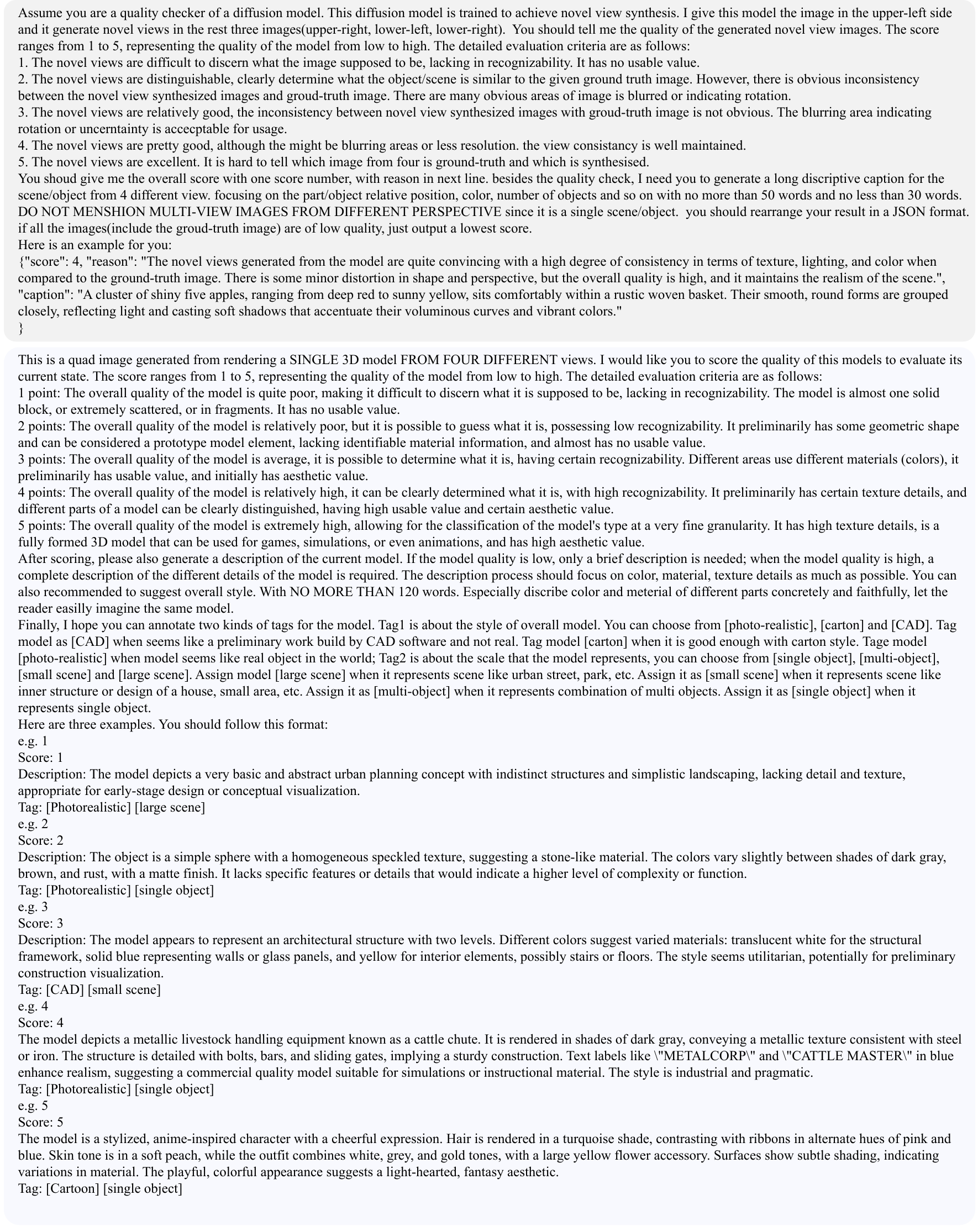}
   \caption{\textbf{Prompt for GPT-4V to generate caption and estimate quality of multi-view images from SV3D~\citep{voleti2024sv3d}, zero123++~\citep{shi2023zero123++} and Objaverse~\citep{deitke2023objaverse}.}} 
   \label{fig:prompt_gpt_check}
\end{figure*}
Detailed prompts are shown in Fig.\ref{fig:prompt_gpt_check}.

\newpage

\subsubsection{Prompts for MV-LLaVA Instruct Tuning}
\label{sup:llava_prompt_detail}
\begin{table}[htbp]
  \centering
  \caption{\textbf{Instruct tuning prompt for SV3D~\citep{voleti2024sv3d} and Zero123++~\citep{shi2023zero123++} multi-view images}}
    \begin{tabular}{p{3cm}|p{10cm}}
    \toprule
    \centering \textbf{prompt type} & \centering \textbf{prompt} \tabularnewline
    \midrule
    \multirow{3}[0]{*}{generate caption} & \textless image\textgreater \textless image\textgreater \textless image\textgreater \textless image\textgreater \textbackslash{}nWhat is this multi-view photo about? generate a short caption for me. \\
          & \textless image\textgreater \textless image\textgreater \textless image\textgreater \textless image\textgreater \textbackslash{}nGenerate a short caption of the following multi-view image. \\
          & \textless image\textgreater \textless image\textgreater \textless image\textgreater \textless image\textgreater \textbackslash{}nCan you describe the main features of this multi-view image for me by a short caption? \\
    \midrule
    \multirow{3}[0]{*}{reasoning} & How about the view consistency of this synthesized multi-view image? \\
          & Do some comments about the view consistency of this synthesized multi-view image. \\
          & What do you think about the view consistency of this synthesized multi-view image? \\
    \midrule
    quality estimation & What do you think about the overall quality of view consistency of three synthesized novel views? Choosing from "poor", "relatively poor", "boardline", "relatively good", "good", "perfect". \\
    \bottomrule
    \end{tabular}%
  \label{tab:instructions}%
\end{table}%

\begin{table}[htbp]
  \centering
  \caption{\textbf{Instruct tuning prompt for Objaverse~\citep{deitke2023objaverse} rendered multi-view images}}
    \begin{tabular}{p{3cm}|p{10cm}}
    \toprule
    \centering \textbf{prompt type} & \centering \textbf{prompt} \tabularnewline
    \midrule
    \multirow{3}[0]{*}{long description} & \textless image\textgreater \textless image\textgreater \textless image\textgreater \textless image\textgreater \textbackslash{}nWhat is this multi-view photo about? generate a long descriptive caption for me. \\
          & \textless image\textgreater \textless image\textgreater \textless image\textgreater \textless image\textgreater \textbackslash{}nGenerate a long descriptive caption of the following multi-view image. \\
          & \textless image\textgreater \textless image\textgreater \textless image\textgreater \textless image\textgreater \textbackslash{}nCan you describe the main features of this multi-view image for me by a long descriptive caption caption? \\
    \midrule
    \multirow{3}[0]{*}{caption} & \textless image\textgreater \textless image\textgreater \textless image\textgreater \textless image\textgreater \textbackslash{}nWhat is this multi-view photo about? generate a short caption for me. \\
          & \textless image\textgreater \textless image\textgreater \textless image\textgreater \textless image\textgreater \textbackslash{}nGenerate a short caption of the following multi-view image. \\
          & \textless image\textgreater \textless image\textgreater \textless image\textgreater \textless image\textgreater \textbackslash{}nCan you describe the main features of this multi-view image for me by a short caption? \\
    \midrule
    quality estimation & What do you think about the overall quality of this 3D model? Choosing from "poor", "relatively poor", "boardline", "relatively good", "good", "perfect". \\
    \midrule
    scale tag & What do you think about the scale of the 3D model represents? Choosing from "single\_object", "multi-object", "small\_scene", "large\_scene". \\
    \midrule
    style tag & What do you think about the overall style of the 3D model? Choosing from "CAD", "Cartoon", "Photo\_realistic". \\
    \bottomrule
    \end{tabular}%
  \label{instructions_obj}%
\end{table}%

\begin{figure*}[h]
  \centering
  \includegraphics[width=1.0\linewidth]{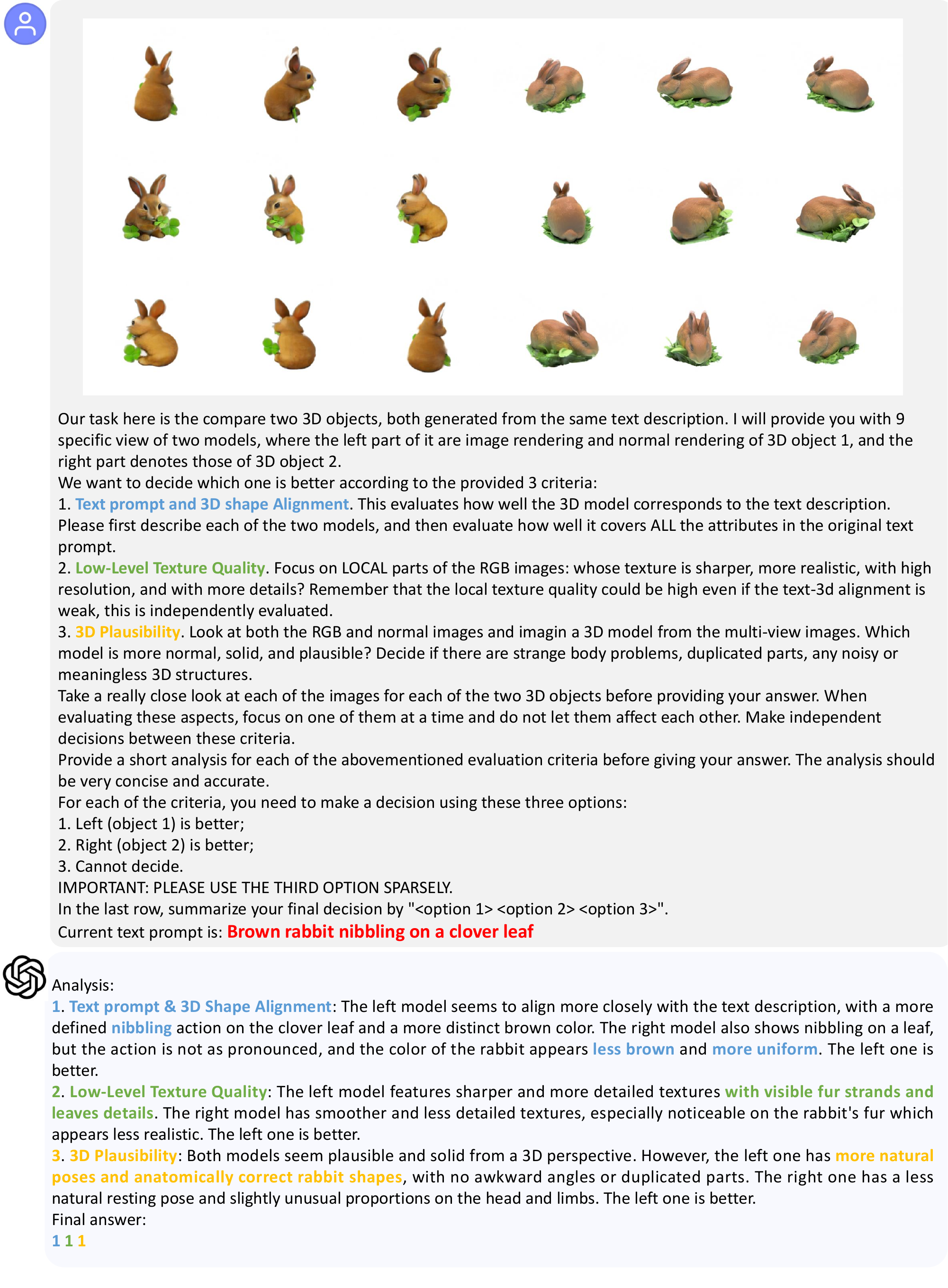}
   \caption{\textbf{A test conversation with GPT-4V~\citep{2023GPT4VisionSC}} of evaluating generated objects.}
   \label{fig:gpt_eval}
\end{figure*}

\subsection{GPT-4V based 3D Object Generation Evaluation.}
\label{sup:gpt_evaluation}
We adopt method proposed in GPTeval3D~\citep{wu2024gpt} for more thorough and human-aligned evaluation of the quality of generated object by different methods. A full test case is shown in Fig.~\ref{fig:gpt_eval}. Left 9-view image is rendered from object generated by \methodname and the right one generated by Instant3D~\citep{li2023instant3d}. We ask GPT-4V to mainly evaluate through comparison based on three dimensions: text-image alignment, low-level texture quality and 3D plausibility. The answer of GPT-4V shows its in depth perception ability of given reasonable comparison well aligned with human preference. We thus choose to use GPT-4V rather than human volunteers to give reasonable evaluation.

We adopt the 110 test prompts proposed in GPTeval3D~\citep{wu2024gpt} to test \methodname generated object comparing with Instant3D~\citep{li2023instant3d}, Zero123++~\citep{shi2023zero123++} and MVDream~\citep{shi2023mvdream}. For each methods, we conditioned model based on 110 test prompts with 4 different seeds, with each methods generates 440 objects, we make 1-to-1 comparison following aforementioned test setting. Results are reported in Tab.~\ref{tab:gpt_eval}. Except MVDream~\citep{shi2023mvdream} (SDS) (which generates single object consuming 30 mins while \methodname only need 5 seconds.). \methodname excels in all three evaluation dimensions, which proves the ability of \methodname in creating high quality 3D objects.

% \begin{table}[htbp]
%   \centering
%     \begin{tabular}{lccc}
%     \toprule
%           & Image-text Alignment & Texture Quality & 3D Plausibility \\
%     \midrule
%     Compared to Instant3D~\citep{li2023instant3d} & 247/116 & 202/162 & 259/110 \\
%     Compared to Zero123++~\citep{shi2023zero123++} & 192/143 & 210/161 & 231/139 \\
%     \bottomrule
%     \end{tabular}%
%     \caption{GPT-4V based evaluation result.}
%   \label{tab:gpt_eval}%
% \end{table}%

\begin{table}[htbp]
  \centering
  \caption{\textbf{GPT-4V based evaluation result.} the result is in format of "number of objects preferred geneated by \methodname / that of other methods". Cases when GPT cannot answer the question or generates "cannot decide" answer are excluded.}
  \scalebox{0.88}{
    \begin{tabular}{lccc}
    \toprule
          & Image-text alignment & Texture quality & 3D plausibility \\
    \midrule
    Compared to Instant3D~\citep{li2023instant3d} (unofficial) & 247 / 116 & 202 / 162 & 259 / 110 \\
    Compared to Zero123++~\citep{shi2023zero123++} & 192 / 143 & 210 / 161 & 231 / 139 \\
    Compared to MVDream~\citep{shi2023mvdream} (GRM) & 290 / 71 & 245 / 131 & 284 / 102 \\
    Compared to MVDream~\citep{shi2023mvdream} (SDS) & 188 / 155 & 173 / 190 & 192 / 150 \\
    \bottomrule
    \end{tabular}%
    }
  \label{tab:gpt_eval}%
\end{table}%

\subsection{Improving Direct 3D Generative Models}

\begin{figure*}[h]
    \centering
    \includegraphics[width=\linewidth]{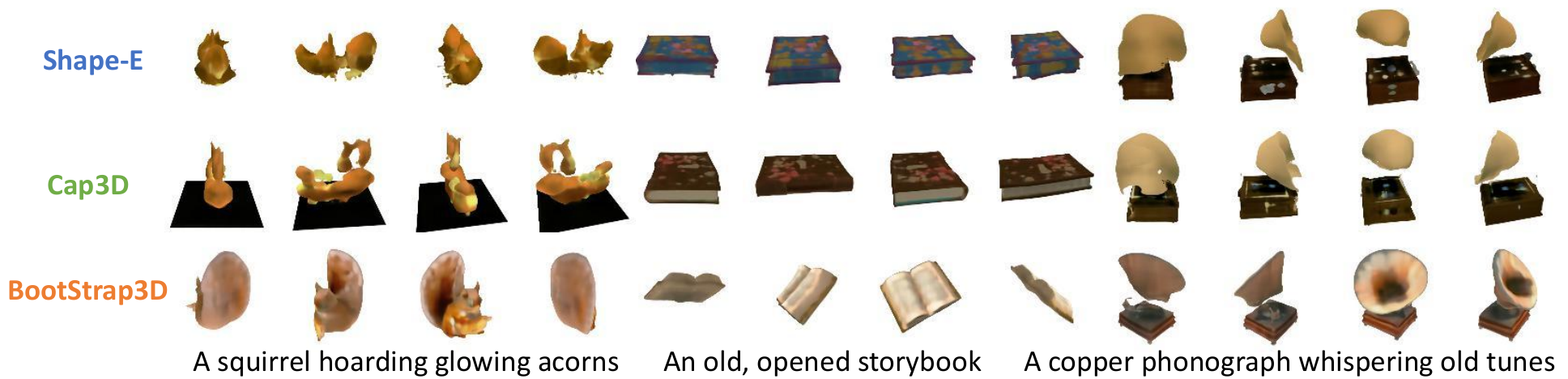}
    \caption{\textbf{Fine tuned Shape-E generation results} that shows better object-text alignment than original Shape-E~\citep{jun2023shap} and finetuned version in Cap3D~\citep{luo2024scalable}.}
    \label{fig_point_e}
\end{figure*}

\begin{table}[h]
    \centering
    \caption{\textbf{Test results on Shape-E.} More accurate and descriptive 3D caption help model to achieve better object-text alignment.} 
    \begin{tabular}{lccc}
    \toprule
    Method & FID $\downarrow$ & CLIP score $\uparrow$ & CLIP-R-precision $\uparrow$ \\ 
    \midrule
    Shape-E & 37.2 & 80.4 & 20.3 \\ 
    Cap3D & 35.5 & 79.1 & 20.0 \\ 
    Ours & 35.3 & 81.2 & 22.1 \\
    \bottomrule
    \label{table:finetune_on_shape_e}
    \end{tabular}
\end{table}

In addition to fine-tuning the multiview diffusion model, we also evaluate our framework on direct 3D generative models, circumventing the use of multi-view images as intermediaries. For this purpose, we selected the Shape-E~\citep{jun2023shap} model for experiment and assess the outcomes following the testing method the same to Cap3D~\citep{luo2024scalable}. Specifically, we fine-tune Shape-E using 250K BS-Objaverse data, ensuring that all entries scored greater than 3, accompanied by more precise and descriptive captions. The metrics for training and testing are consistent with those employed in Cap3D~\citep{luo2024scalable}. Some qualitative results are presented in Fig.\ref{fig_point_e}, where our finetuned verson can generate object that follow text prompt more precisely. Quantitative results are detailed in Tab.\ref{table:finetune_on_shape_e}, where more accurate and desciptive captions than Cap3D can significantly improve metrics like CLIP score. Our findings indicate that improved data quality can significantly enhance object-text alignment and visual quality of Shape-E. This experiment substantiates that our pipeline, characterized by detailed captions and quality filtering, is also effective for direct 3D objects generation represented by neural field.

\subsection{More Results Visualization}
\subsubsection{Comparison with Other Methods}
\label{sup_more_com}
\begin{figure*}[h]
  \centering
  \includegraphics[width=1.0\linewidth]{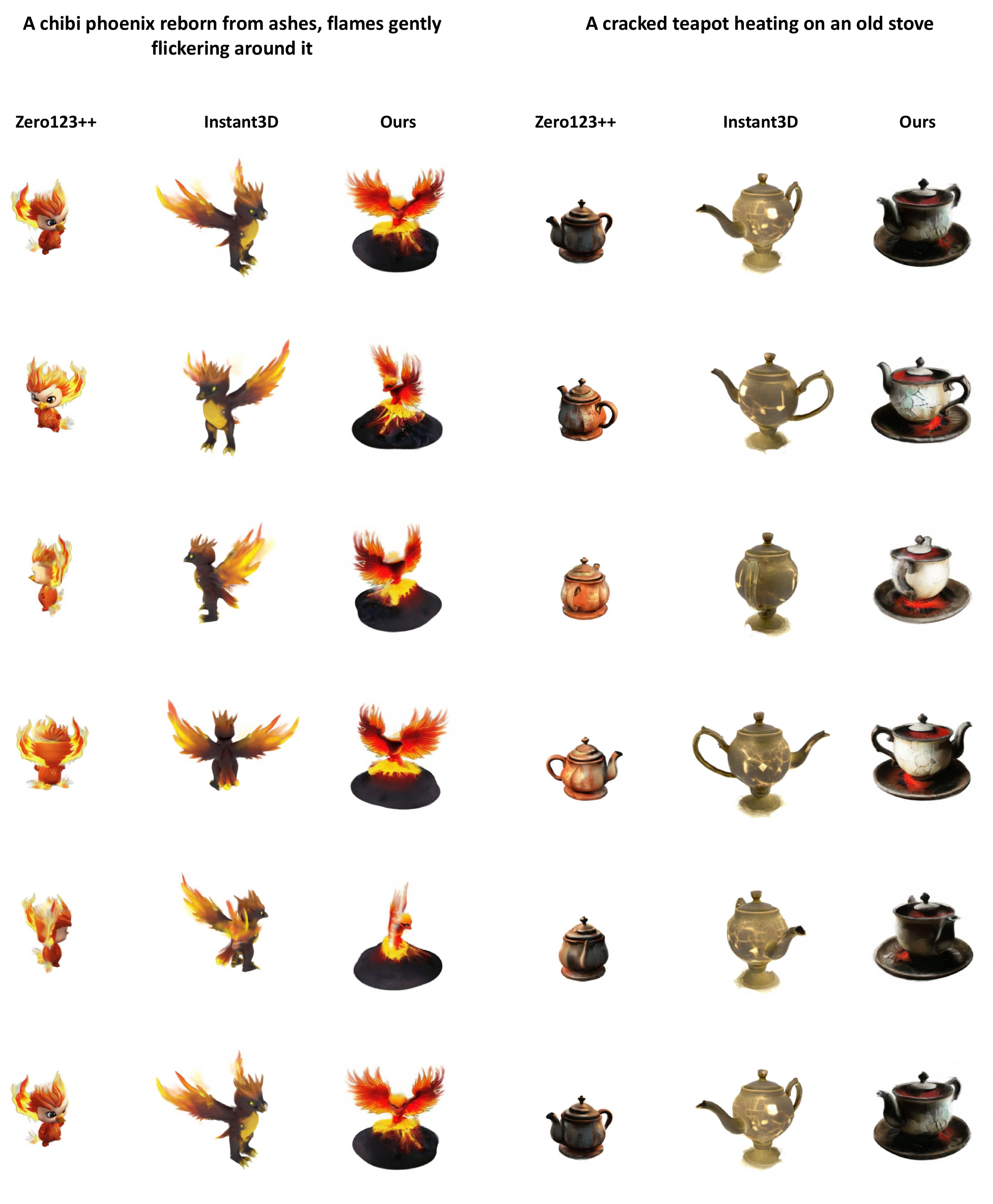}
   \caption{\textbf{Visualization of generated objects compared to other edge-cutting methods}} 
   \label{fig:more_compare_1}
\end{figure*}

\begin{figure*}[h]
  \centering
  \includegraphics[width=1.0\linewidth]{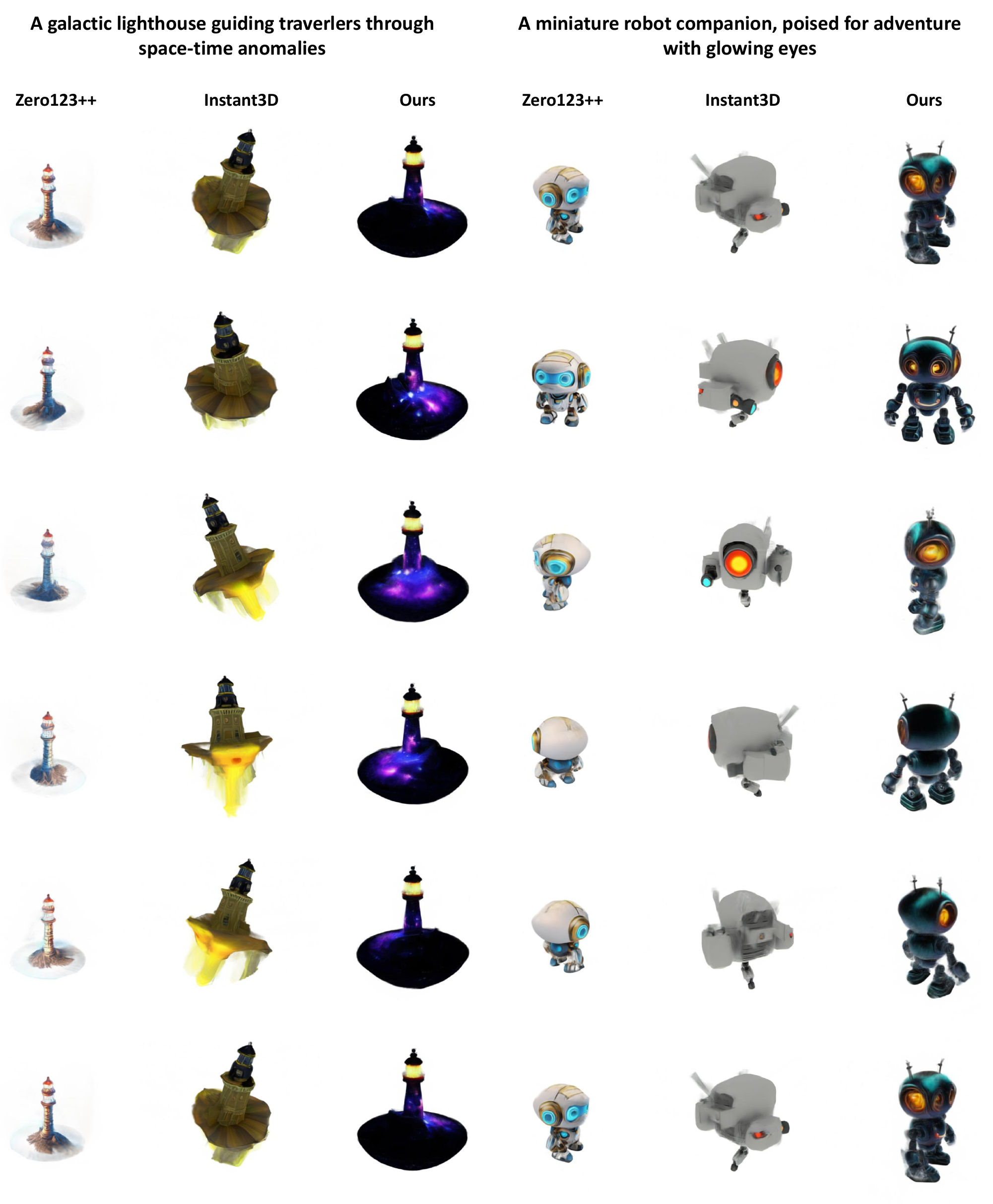}
   \caption{\textbf{Visualization of generated objects compared to other edge-cutting methods}} 
   \label{fig:more_compare_2}
\end{figure*}

\begin{figure*}[h]
  \centering
  \includegraphics[width=1.0\linewidth]{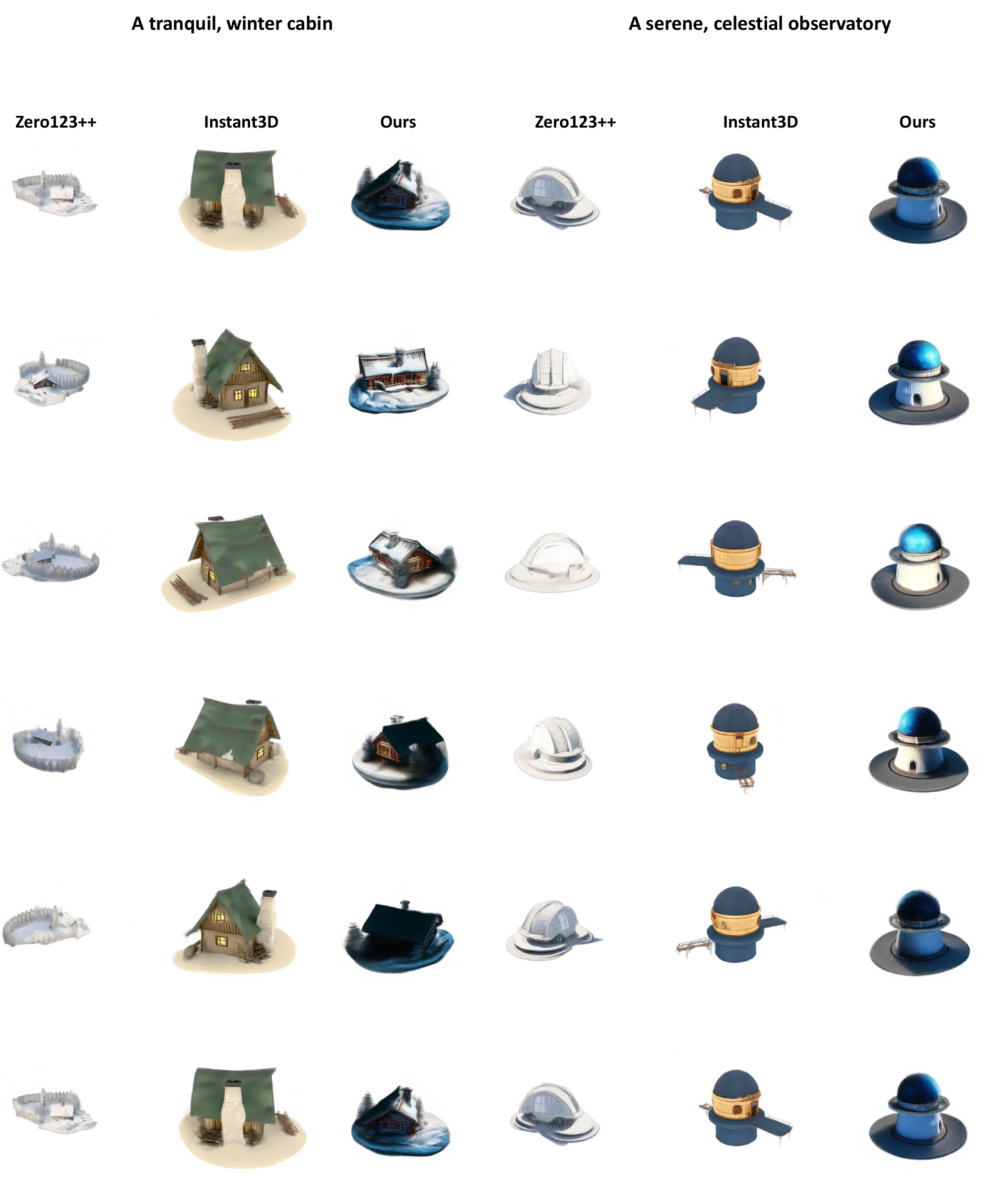}
   \caption{\textbf{Visualization of generated objects compared to other edge-cutting methods}} 
   \label{fig:more_compare_3}
\end{figure*}

\begin{figure*}[h]
  \centering
  \includegraphics[width=1.0\linewidth]{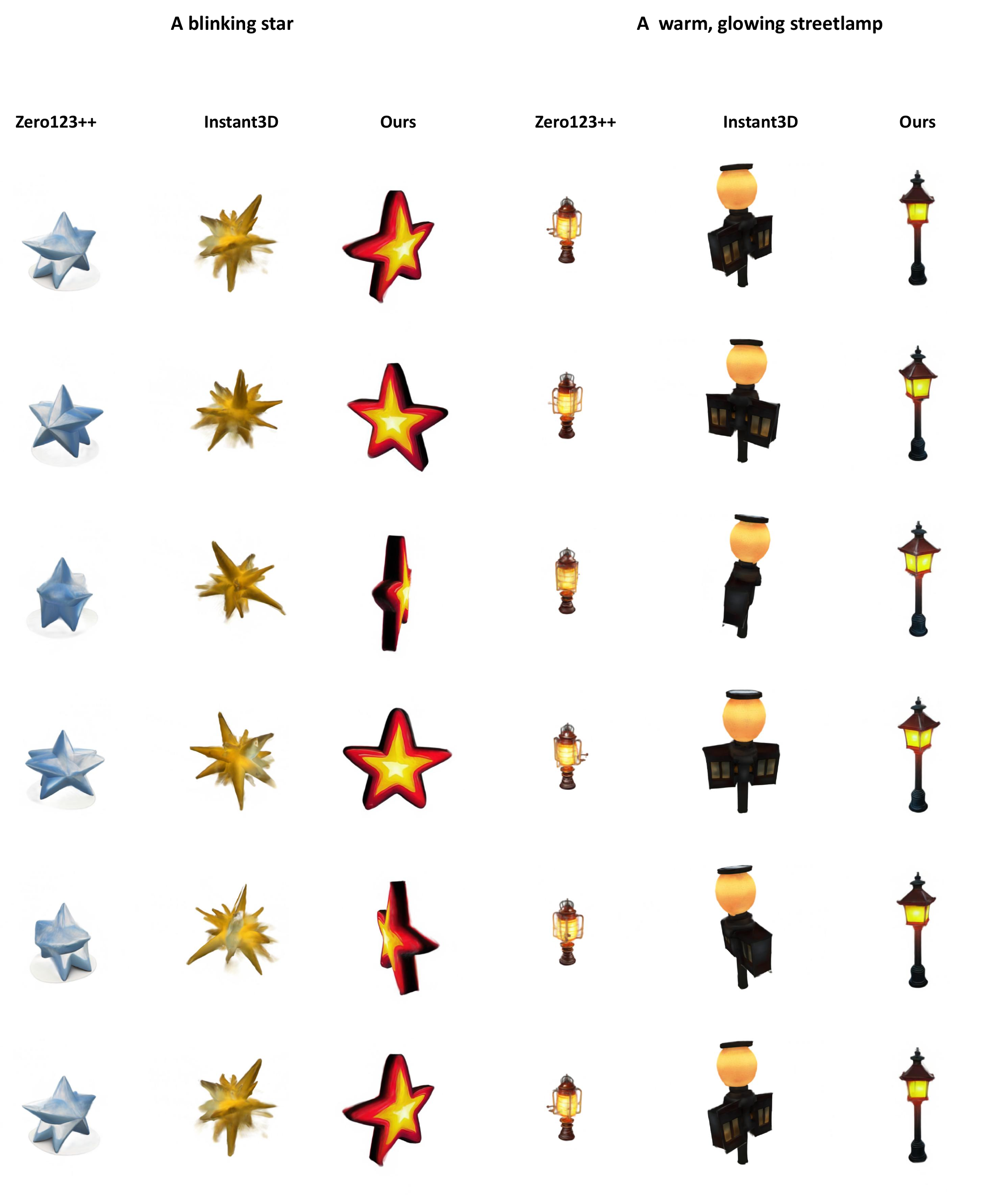}
   \caption{\textbf{Visualization of generated objects compared to other edge-cutting methods}} 
   \label{fig:more_compare_4}
\end{figure*}

\begin{figure*}[h]
  \centering
  \includegraphics[width=1.0\linewidth]{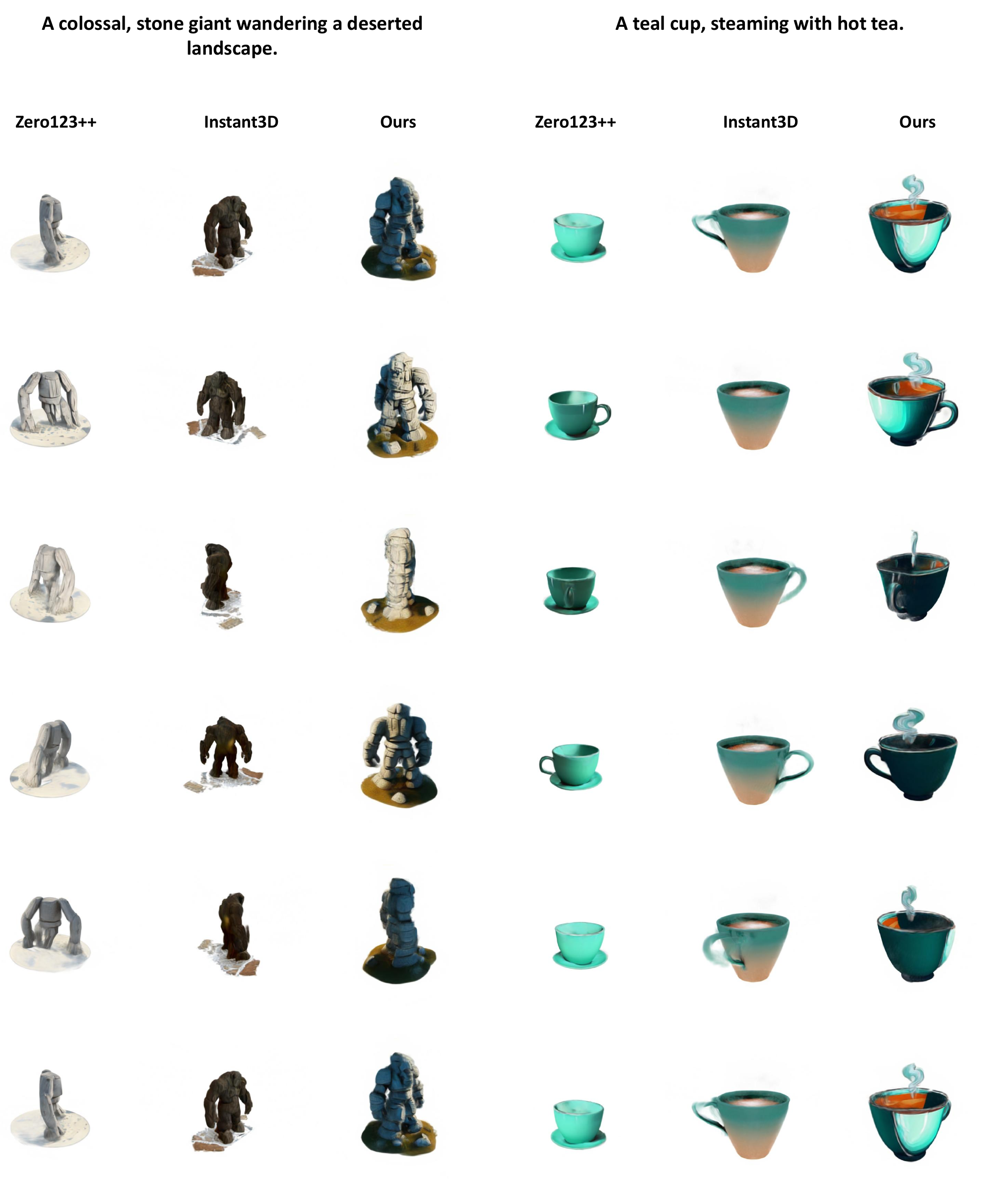}
   \caption{\textbf{Visualization of generated objects compared to other edge-cutting methods}} 
   \label{fig:more_compare_5}
\end{figure*}

\subsubsection{Visualization of Generated objects with Different Styles}

\label{sup_more_style}
\begin{figure*}[h]
  \centering
  \includegraphics[width=1.0\linewidth]{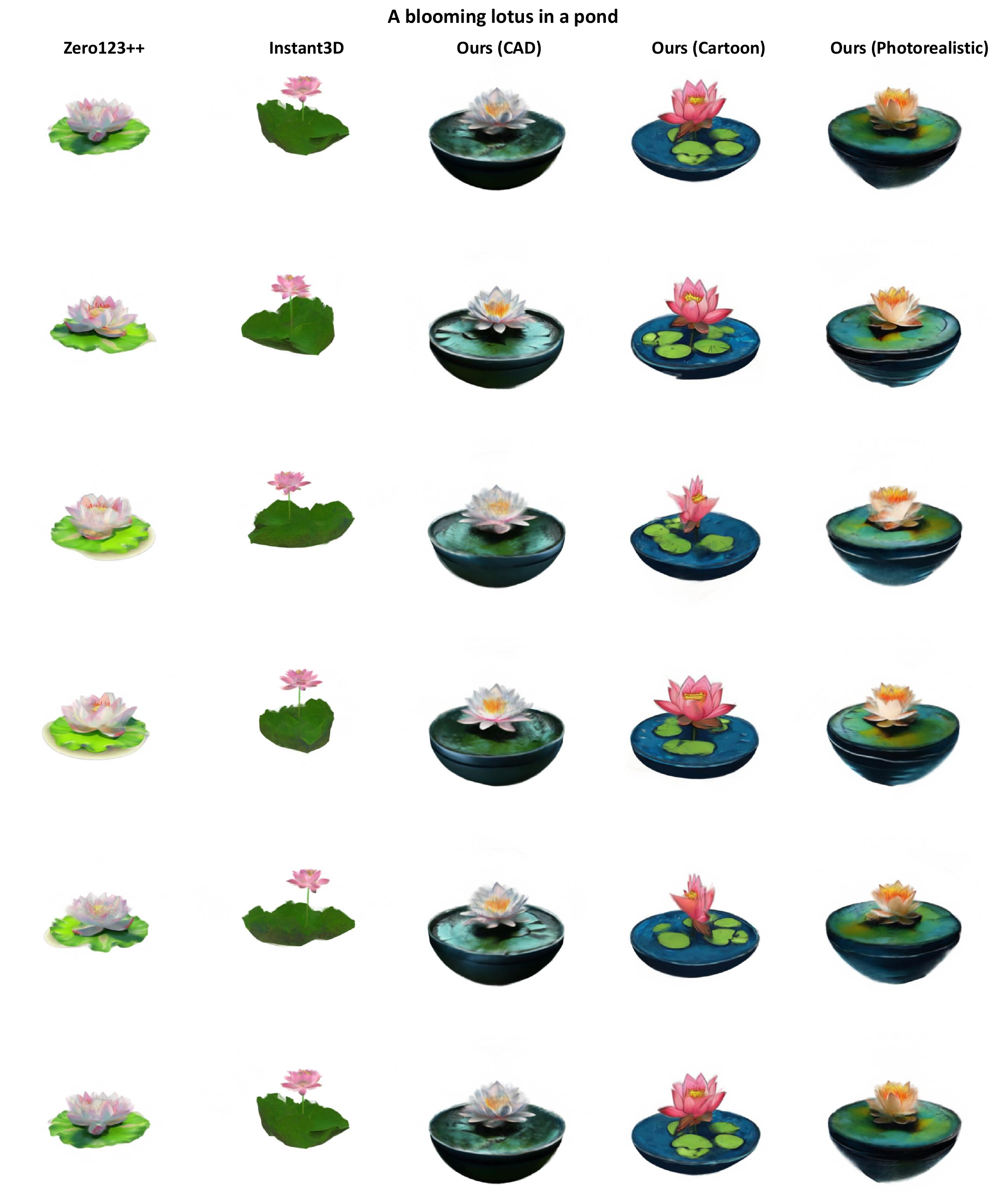}
   \caption{\textbf{Visualization of generated objects compared to other edge-cutting methods with different style control.}} 
   \label{fig:more_style_1}
\end{figure*}

\begin{figure*}[h]
  \centering
  \includegraphics[width=1.0\linewidth]{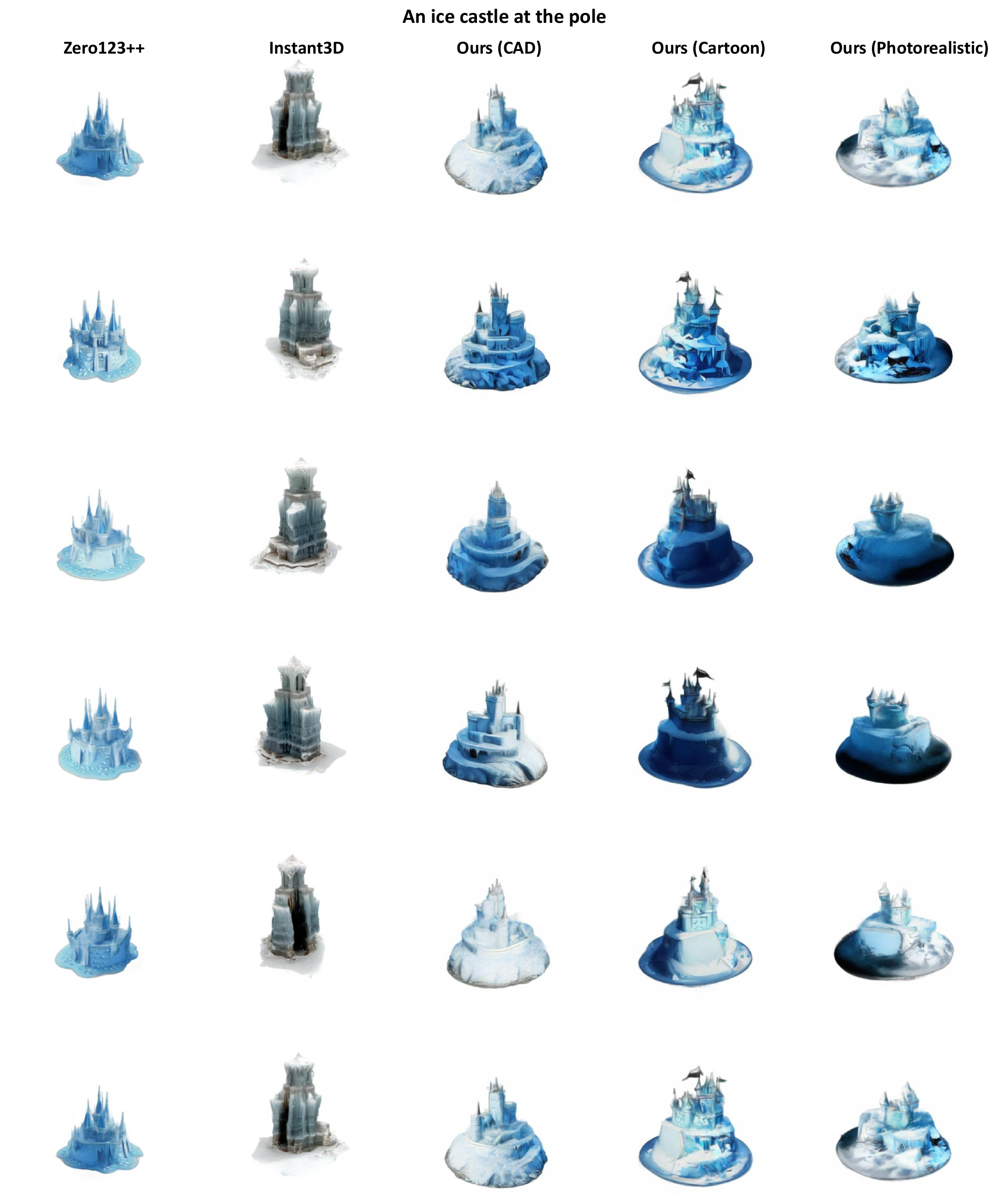}
   \caption{\textbf{Visualization of generated objects compared to other edge-cutting methods with different style control.}} 
   \label{fig:more_style_2}
\end{figure*}

\begin{figure*}[h]
  \centering
  \includegraphics[width=1.0\linewidth]{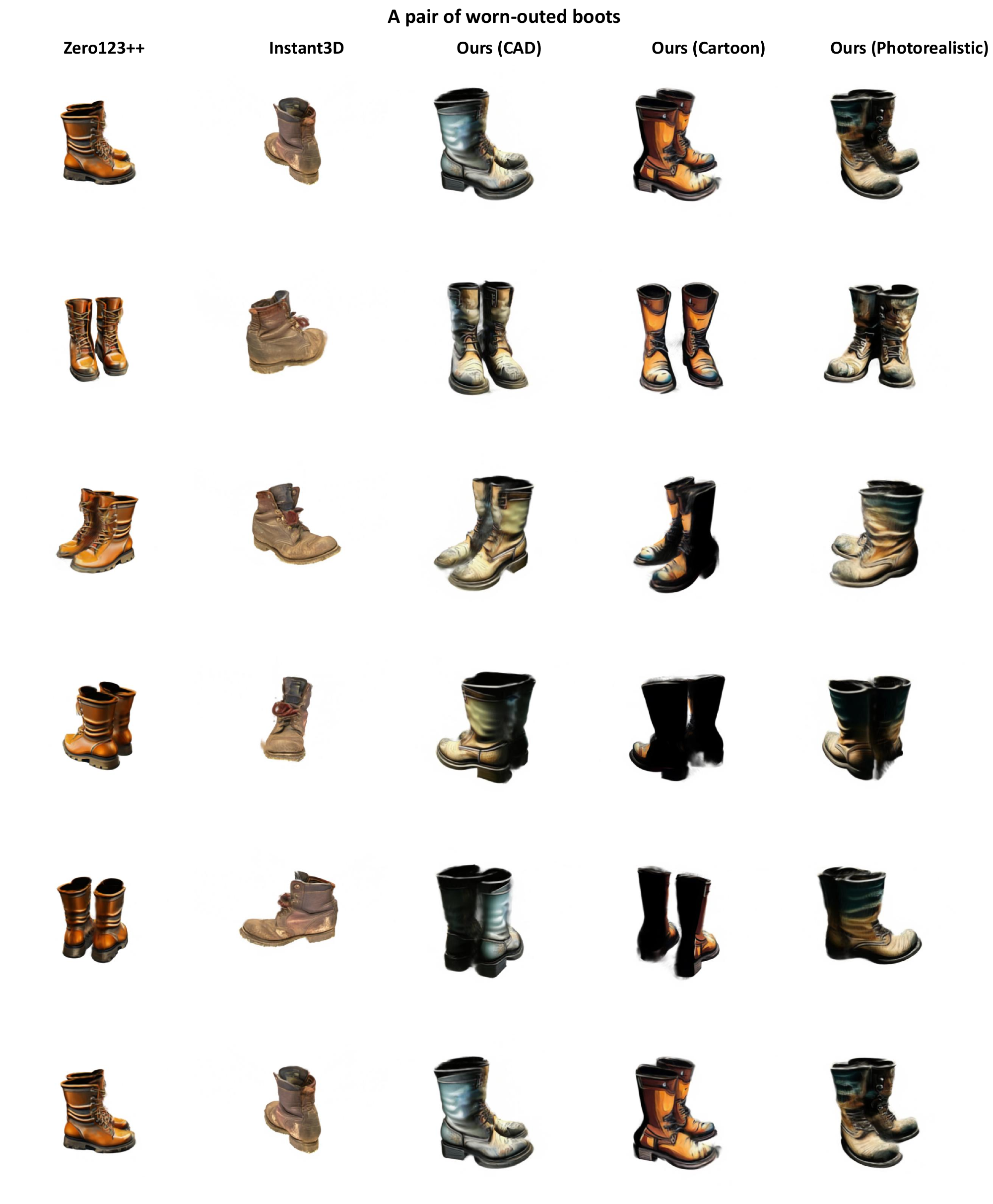}
   \caption{\textbf{Visualization of generated objects compared to other edge-cutting methods with different style control.}} 
   \label{fig:more_style_3}
\end{figure*}

% \begin{figure*}[h]
%   \centering
%   \includegraphics[width=1.0\linewidth]{figures/more_style_4.pdf}
%    \caption{\textbf{Visualization of generated objects compared to other edge-cutting methods with different style control.}} 
%    \label{fig:more_style_4}
% \end{figure*}

\begin{figure*}[h]
  \centering
  \includegraphics[width=1.0\linewidth]{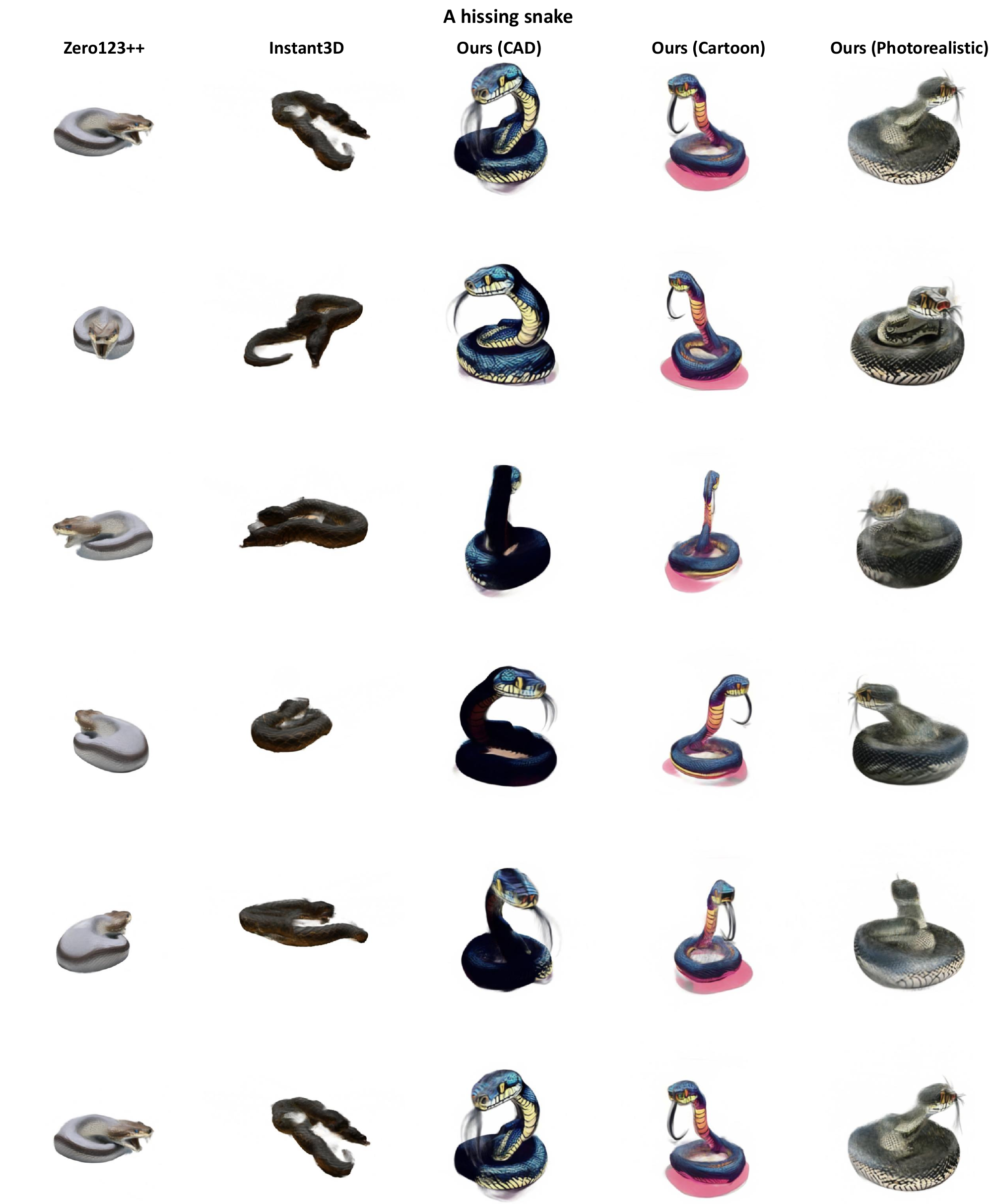}
   \caption{\textbf{Visualization of generated objects compared to other edge-cutting methods with different style control.}} 
   \label{fig:more_style_5}
\end{figure*}

\begin{figure*}[h]
  \centering
  \includegraphics[width=1.0\linewidth]{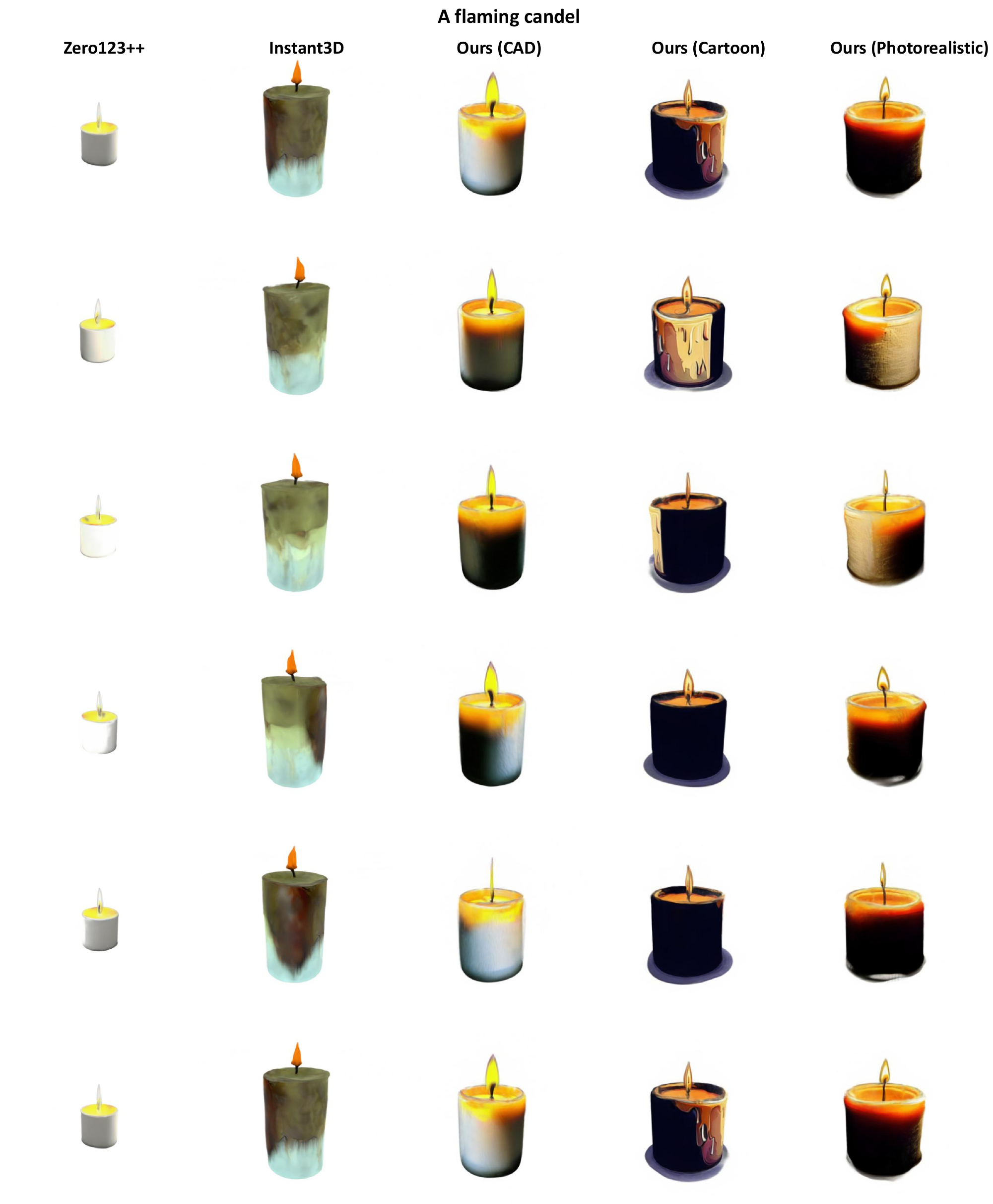}
   \caption{\textbf{Visualization of generated objects compared to other edge-cutting methods with different style control.}} 
   \label{fig:more_style_6}
\end{figure*}

\subsection{Broader Impacts}
\label{sup:broader_impacts}
\textbf{Potential positive societal impacts:} The proposed framework, Bootstrap3D, enhances the quality and consistency of 3D models, which can benefit various industries such as entertainment, education, virtual reality, and digital art. By generating and sharing a large synthetic dataset of high-quality synthetic multi-view images, We will promotes open access to resources that can accelerate progress in the field. The model and data can serve as educational tools for students and researchers, fostering learning and innovation in machine learning and 3D modeling. 

\textbf{Potential negative societal impacts:} High-quality 3D models could be used to create deepfakes or misleading content, which may contribute to disinformation or malicious activities. Monitoring and Defense Mechanisms: Developing tools to detect and prevent the misuse of the generated 3D models, particularly in contexts like disinformation and surveillance. There may be unintended biases in the generated data or models, leading to unfair treatment of specific groups if the technology is deployed in applications affecting societal decision-making.

\end{document}